\documentclass[lettersize,journal]{IEEEtran}
\usepackage{amsmath,amsfonts}
\usepackage{algorithmic}
\usepackage{algorithm}
\usepackage{array}
\usepackage[caption=false,font=footnotesize,labelfont=rm,textfont=rm]{subfig}
\usepackage{textcomp}
\usepackage{stfloats}
\usepackage{url}
\usepackage{verbatim}
\usepackage{graphicx}
\usepackage{cite}
\usepackage{multirow}
\usepackage{color}
\usepackage{xcolor}
\DeclareSubrefFormat{myparens}{#1(#2)}
\usepackage{makecell}

\begin{document}

\title{ARS-DETR: Aspect Ratio-Sensitive Detection Transformer for Aerial Oriented Object Detection}

\author{Ying Zeng$^{1}$, Yushi Chen$^{1}$, \textit{Member, IEEE}, Xue Yang$^{2}$, Qingyun Li$^{1}$, Junchi Yan$^{2}$, \textit{Senior Member, IEEE}\\
\thanks{
This work was supported by the Natural Science Foundation of China under Grant 62371169, 61971164 and U20B2041 (Corresponding author: Yushi Chen.)

Ying Zeng, Yushi Chen and Qingyun Li are with the School of Electronics and Information Engineering, Harbin Institute of Technology, Harbin 150001, China (e-mail: 22S005064@stu.hit.edu.cn; chenyushi@hit.edu.cn; 21B905003@stu.hit.edu.cn)

Xue Yang is with OpenGVLab, Shanghai AI Laboratory, Shanghai 200030, China (e-mail: yangxue-2019sjtu@sjtu.edu.cn)

Junchi Yan is with School of Electronic Information and Electrical Engineering, and MoE Key Lab of Artificial Intelligence, Shanghai Jiao Tong University, Shanghai 200030, China (e-mail: yanjunchi@sjtu.edu.cn)

}
\thanks{
}}

\markboth{Journal of \LaTeX\ Class Files,~Vol.~14, No.~8, August~2021}%
{Shell \MakeLowercase{\textit{et al.}}: A Sample Article Using IEEEtran.cls for IEEE Journals}


\maketitle

\begin{abstract}
    Existing oriented object detection in aerial images has progressed a lot in recent years and achieved a favorable success. However, high-precision oriented object detection in aerial images remains a challenging task. 
    Some recent works have adopted the classification-based method to predict the angle in order to address boundary problem in angle.
    However, we have found that these works often neglect the sensitivity of objects with different aspect ratios to angle.
    At the same time, it is worth exploring a suitable way to improve the emerging transformer-based approaches 
    in order to adapt them to oriented object detection.
    In this paper, we propose an \textbf{A}spect \textbf{R}atio \textbf{S}ensitive \textbf{DE}tection \textbf{TR}ansformer, termed ARS-DETR, for oriented object detection in aerial images. Specifically, a new angle classification method, called Aspect Ratio aware Circle Smooth Label (AR-CSL), is proposed to smooth the angle label in a more reasonable way and discard the hyperparameter that introduced by previous work (e.g. CSL). Then, a rotated deformable attention module is designed to rotate the sampling points with the corresponding angles and eliminate the misalignment between region features and sampling points. Moreover, a dynamic weight coefficient according to the aspect ratio is adopted to calculate the angle loss. Comprehensive experiments on several challenging datasets demonstrate that our method achieves a competitive performance in the high-precision oriented object detection task.
\end{abstract}


\section{Introduction}
Object detection in aerial images 
has always been
a hot spot in the remote sensing community. 
With the rapid increase of a large number of available high-resolution aerial images\cite{liu2017high, xia2018dota, cheng2022anchor}, accurately and effectively detection in these aerial images has become a crucial issue.

Benefiting from the development of deep learning, the emergence of deep Convolutional Neural Networks (CNNs) greatly influenced the design of detectors and has achieved a favorable performance in generic object detection.
Instead of using handcrafted features for detection, which is cumbersome and not accurate enough, CNNs could learn from the training data and update themselves iteratively, 
exhibiting a strong ability to extract high-level and robust features for accurate detection.
Numerous advanced detectors have also been proposed to detect the objects by using horizontal bounding boxes (HBBs).

Compared with generic images, objects in aerial images often exhibit a wide variety of scales, aspect ratios, and orientations, and sometimes they are arranged densely.
When simply using HBBs to detect these objects, HBBs cannot fit these objects very well and thus will include abundant background or overlap with other objects. Therefore, oriented object detection, which adopts the oriented bounding boxes (OBBs) to represent the objects, is more suitable for aerial object detection.

\begin{figure}[!tb]
    \centering
    \subfloat{
        \begin{minipage}[h]{0.45\linewidth}
            \centering
            \includegraphics[width=0.98\linewidth]{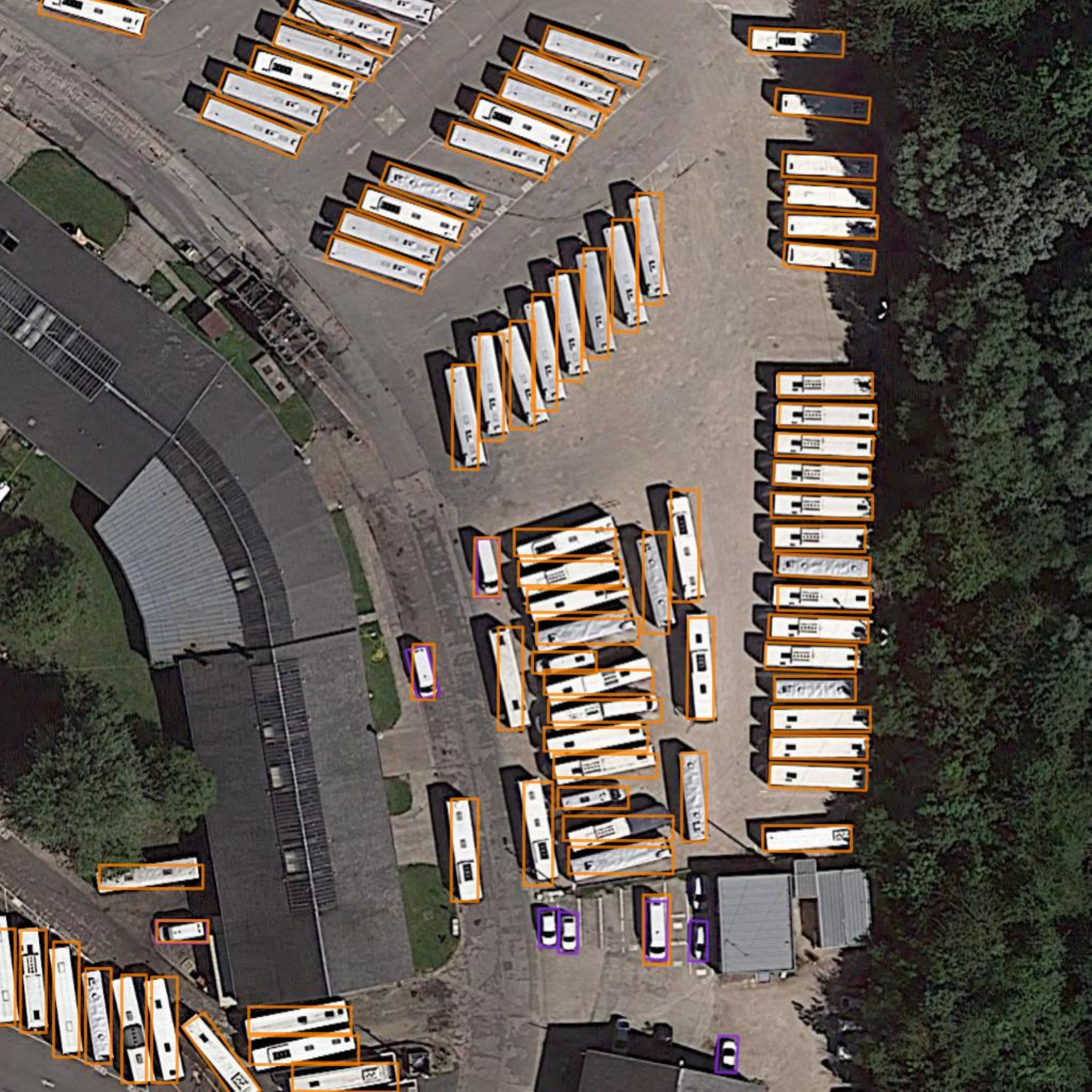}
        \end{minipage}%
        \label{fig:bad1}
    }
    \subfloat{
        \begin{minipage}[h]{0.45\linewidth}
            \centering
            \includegraphics[width=0.98\linewidth]{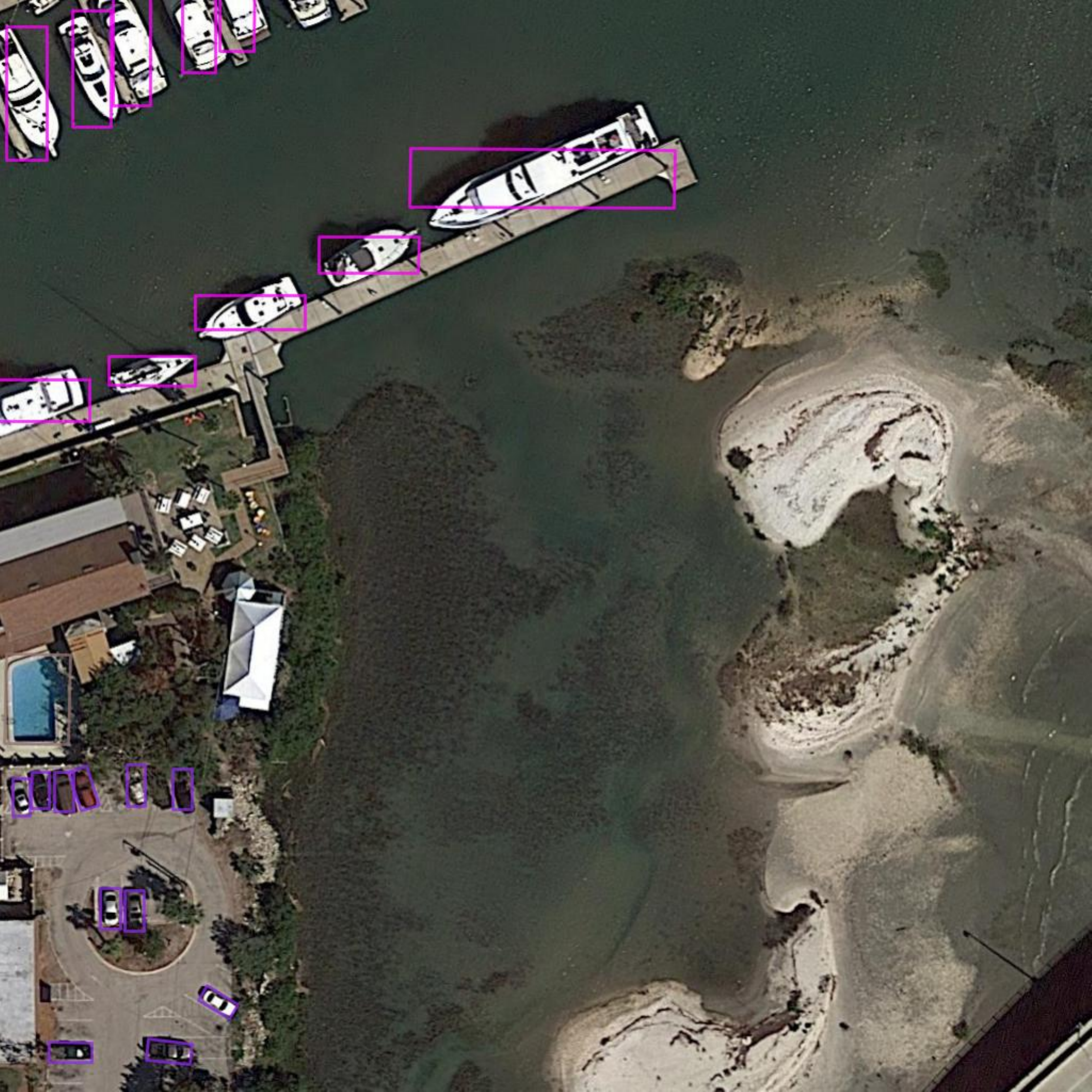}
        \end{minipage}%
        \label{fig:bad2}
    }\\
    \subfloat{
        \begin{minipage}[h]{0.45\linewidth}
            \centering
            \includegraphics[width=0.98\linewidth]{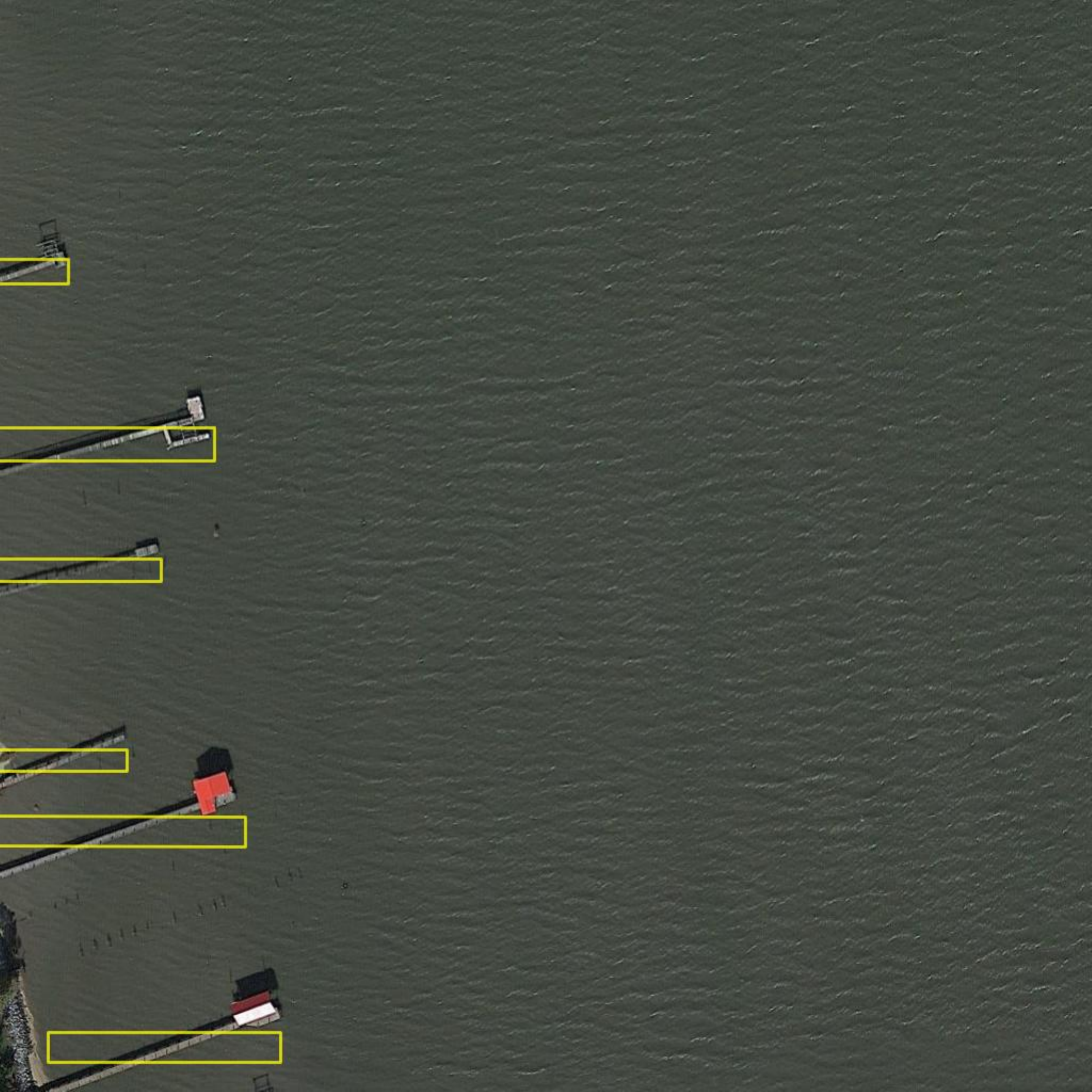}
        \end{minipage}
        \label{fig:bad3}
    }
    \subfloat{
        \begin{minipage}[h]{0.45\linewidth}
                \centering
                \includegraphics[width=0.98\linewidth]{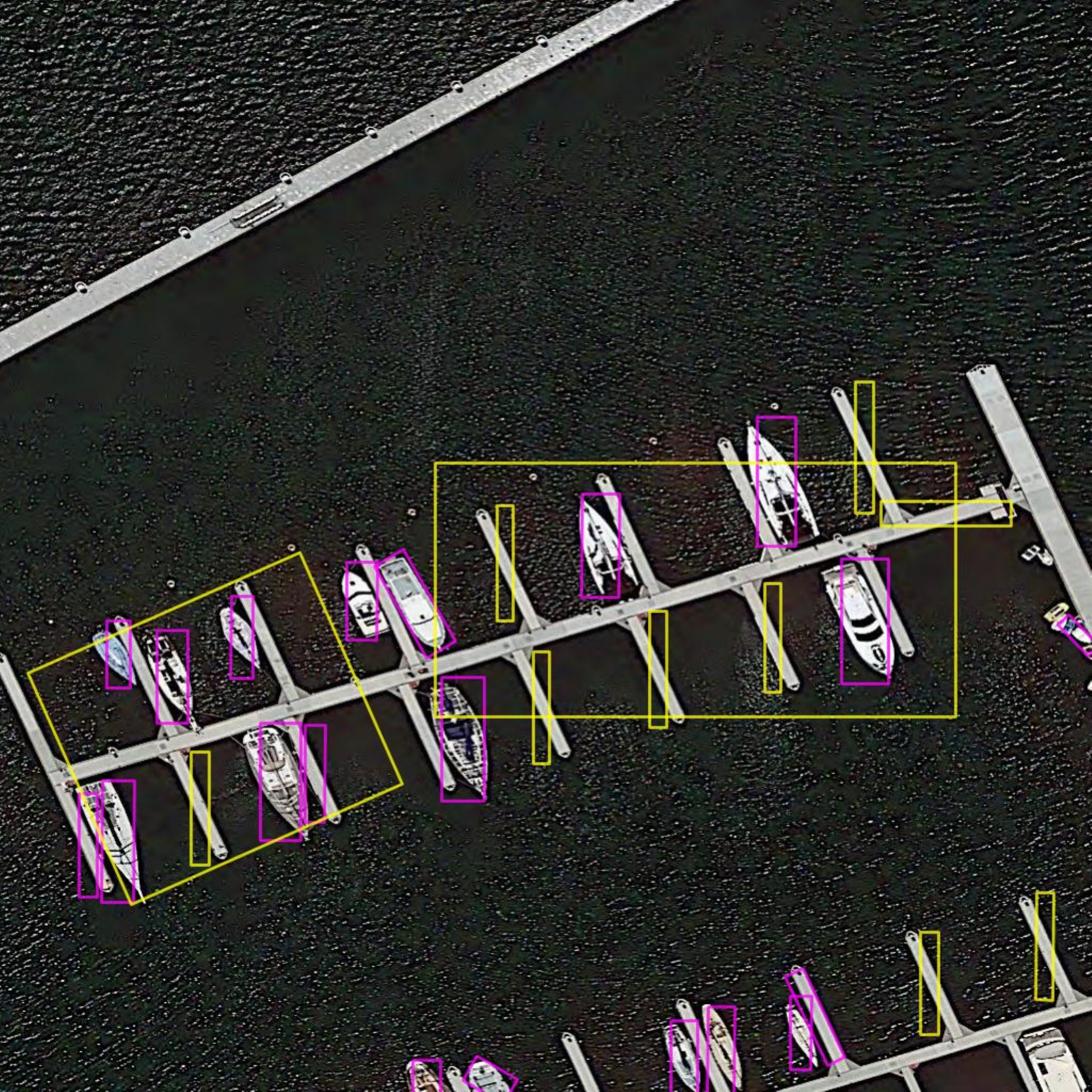}
            \end{minipage}
            \label{fig:bad4}
    }
    \vspace{8pt}
    \caption{Even though the angle prediction is inaccurate, it still obtains a high performance in terms of AP$_{50}$.}
    \label{fig:inaccurate_angle_prediction}
    \vspace{-8pt}
\end{figure}

As a recently emerged task in remote sensing, oriented object detection exhibits a strong ability in analysing the objects in aerial images and many advanced oriented object detectors have been proposed and achieved a favorable performance in aerial images\cite{ding2019learning, yang2019scrdet, xu2020gliding, li2022oriented}.
However, numerous detectors treat oriented object detection as generic detection involving an angle that needs to be predicted. To achieve this, they simply introduce an additional angle parameter in the detection head. Consequently, their angle predictions are not very accurate, as shown in Fig.~\ref{fig:inaccurate_angle_prediction}.
Nevertheless, these detectors can still obtain a fairly good results under current metric i.e. AP$_{50}$, indicating that AP$_{50}$ is not accurate enough to reflect the performance of oriented object detectors and high-precision oriented object detection in aerial images still remains a challenging task.

Angle, as a unique parameter in oriented object detection, plays a vital role in high-precision detection. 
At the same time, the characteristics of angle also make it difficult to predict and 
therefore require more attention.
Firstly, the periodicity of angle will cause the discontinuity at the boundary, leading to a suboptimal optimization during the training process. Secondly, 
objects with different aspect ratios exhibit varying sensitivities to angle,
which is neglected by most of oriented object detectors. 
Objects with a small aspect ratio, especially those resembling squares, exhibit reduced sensitivity to variations in angle.
Consequently, even in situations where there exist notable disparities in the predicted angles, which indicates a significant angle deviation, they are capable of maintaining a high Skew Intersection over Union (SkewIoU) between the 
targets and predictions.
In contrast, even a slight angle deviation can drastically degrade the SkewIoU in cases where objects have a large aspect ratio.

Among a large number of angle prediction methods, classification-based method shows a favorable performance \cite{yang2020arbitrary,yang2021dense,wang2022gaussian}. 
Specifically, it decouples the angle from the bounding boxes and transforms the angle prediction into a classification task, thereby eliminating the boundary problem \cite{yang2021rethinking}.
Moreover, \cite{ming2023task} also shows a strong potential of classification-based method in high-precision oriented object detection. Nevertheless, there still exists some issues, such as ignoring the correlation between the angle and aspect ratio completely, introducing hyperparameter (e.g. window radius in \cite{yang2020arbitrary}), and etc. 
Thus, the accuracy of the angle prediction is hindered to some extent.

Recently, the transformer-based detectors \cite{carion2020end, zhu2021deformable} have revived the object detection task. Without additional complicated hand-designed components like preset Anchor or Non-Maximum Suppression (NMS), DEtection TRansformer (DETR \cite{carion2020end}) regards object detection as a set prediction task and assigns labels by bipartite graph matching, which achieves a comparable performance with classical detectors like Faster RCNN \cite{ren2015faster}. Existing DETR derivatives \cite{zhu2021deformable, meng2021conditional, li2022dn, zhang2023dino, wang2022anchor, liu2022dab} dramatically improve detection performance and convergence speed, exhibiting great potential of Transformer for high-precision object detection. Although some DETR-based oriented object detection methods \cite{ma2021oriented, dai2022ao2} have been proposed, they still use regression to predict angle and do not take into account the issues caused by boundary discontinuity. Meanwhile, they predict angle in a simple way and do not explore how to embed angle information into DETR. How to use DETR more naturally in oriented object detection is still a research topic.

In this paper, we propose an Aspect Ratio Sensitive DEtection TRansformer to achieve oriented object detection in aerial images, called ARS-DETR. Specifically, a hyperparametric free Aspect Ratio aware Circle Smooth Label (AR-CSL) is designed to represent the relationship of adjacent angles according to the aspect ratios of objects. Considering the sensitivity of different objects to angle, AR-CSL uses the SkewIoU of the objects with different aspect ratios under each angle deviation to smooth the angle labels. Then, We also propose a rotated deformable attention module to embed the angle information into detector to align the features. 
Finally, we adopt the aspect ratio sensitive matching and loss strategy to enable dynamic adjustment of the detector's training, thereby reducing the burden of model training.
Extensive experiments on different aerial datasets demonstrate the effectiveness of ARS-DETR in high-precision oriented object detection. In summary, our contributions lie in four-folds as follows:
\begin{itemize}
    \item 
    We analyze the influence of angle deviation in the oriented object detection in detail and give the corresponding formula. Additionally, we also analyze the flaws of the current oriented object detection metric (i.e. AP$_{50}$).
    \item 
    A new angle classification method called AR-CSL is designed to smooth angle labels in a more reasonable way. This method adopts the values of the SkewIoU of objects with different aspect ratios under each angle deviation, while also eliminating the hyperparameter of window radius that was introduced by previous work.
    \item 
    We propose an angle embedded Rotated Deformable Attention module (RDA) to incorporate the angle information for extracting the aligned features. Meanwhile, the Aspect Ratio sensitive Matching (ARM) and Aspect Ratio sensitive Loss (ARL) are developed to adaptively adjust the focus on the angle based on the aspect ratio of the object. In addition, we also combine with DeNoising strategy (DN) to further improve the performance of DETR-based method for oriented object detection.
    \item Extensive experiments on three public aerial datasets: DOTA-v1.0, DIOR-R and OHD-SJTU demonstrate the effectiveness of the proposed model. ARS-DETR achieves a competitive performance on high-precision oriented object detection in the all datasets. 
\end{itemize}

\section{Related Work}\label{sec:related_works}

\subsection{Oriented Object Detection}
As an emerging task, oriented object detection has made great progress in recent years. The simple solution \cite{ma2018arbitrary} for oriented object detection task is to change the Anchor or Region of Interests (RoIs) from the horizontal type to the rotated type. RoI-Transformer \cite{ding2019learning} constructs the geometry transformation to rotate the proposals to locate the objects more accurately. To address the feature misalignment in refined single-stage detectors, R$^3$Det \cite{yang2021r3det} and S$^2$A-Net \cite{han2021align} adopt the feature alignment module to get a more accurate location. However, these mainstream regression-based methods often suffer the boundary problem \cite{yang2020arbitrary} due to the predictions beyond the defined range and need additional complicated treatment. SCRDet \cite{yang2019scrdet} designs a novel IoU-Smooth L1 Loss to alleviate the sudden increase in loss caused by angle periodicity and edge exchangeability, which reduces the difficulty of model training. 

CSL \cite{yang2020arbitrary} transforms the prediction of angle from regression to classification, thereby eliminating the boundary problem. It is achieved through the design of a Circle Smooth Label. Gliding vertex \cite{xu2020gliding} glides the vertex of the horizontal bounding box to accurately represent a multi-oriented object. 
GWD \cite{yang2021rethinking}, KLD \cite{yang2021learning}
and
KFIoU \cite{yang2023kfiou} convert the rotated bounding box into a Gaussian distribution to avoid the boundary discontinuity and square-like issue in oriented object detection.
PSC \cite{yu2023psc} provides a unified framework for various periodic fuzzy issues in oriented object detection by mapping rotational periodicity of different cycles into phase of different frequencies.

\subsection{Angle Classification-based Oriented Object Detection}
The classification-based angle prediction method is a novel and effective approach to circumvent the boundary problem while predicting angle accurately, and it has also made a lot of progress. 
CSL\cite{yang2020arbitrary} discretizes the angle variable into 180 categories and smooths the angle label via a Gaussian window function. 
The CSL method directly promotes the development of classification-based oriented object detection algorithms.
DCL \cite{yang2021dense} uses dense coded label to reduce the amount of computation and parameters of CSL.
\cite{wang2022gaussian} adopts a dynamic weighting mechanism based on CSL to perform precise angle estimation for rotated objects.
To overcome the challenges of ambiguity and high costs in angle representation, 
\cite{wang2022multigrained} proposes a multi-grained angle representation method, consisting of coarse-grained angle classification and fine-grained angle regression.
TIOE \cite{ming2023task} proposes a progressive orientation estimation strategy to approximate the orientation of objects with n-ary codes.
AR-BCL \cite{xiao2023aspect} uses an aspect ratio-based bidirectional coded label to solve the square-like detection issue \cite{yang2021dense}.
In contrast, the new angle encoding technique proposed in this paper is free from hyperparameters, boundary problem, and square-like issue. Furthermore, it explores the potential of angle classification in high-precision detection, which is overlooked by most of the above methods.

\begin{figure*}[!tb]
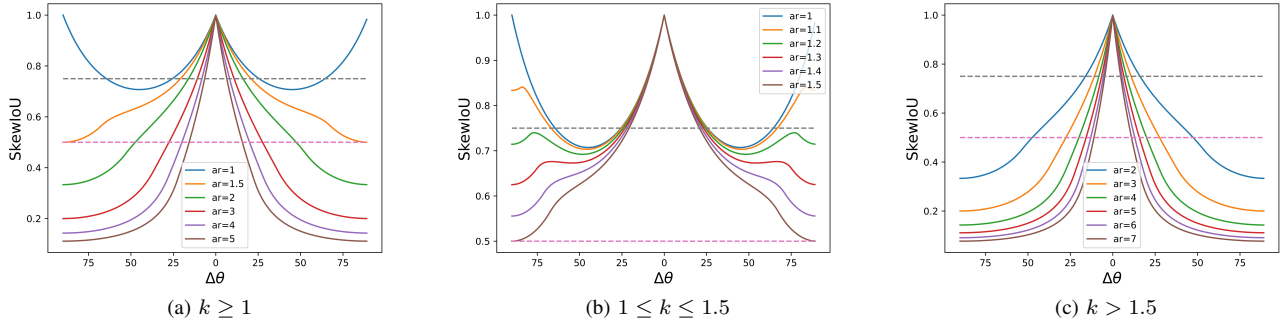

    \begin{center}
        \subfloat[$k \geq 1$]{
            \begin{minipage}[t]{0.315\linewidth}
                \centering
                \includegraphics[width=1.0\linewidth]{figures/SkewIoU1.pdf}
            \end{minipage}%
            \label{fig:skewiou1}
        }
        \subfloat[$1 \leq k \leq 1.5$]{
            \begin{minipage}[t]{0.315\linewidth}
                \centering
                \includegraphics[width=1.0\linewidth]{figures/SkewIoU2.pdf}
            \end{minipage}%
            \label{fig:skewiou2}
        }
        \subfloat[$k > 1.5$]{
            \begin{minipage}[t]{0.315\linewidth}
                \centering
                \includegraphics[width=1.0\linewidth]{figures/SkewIoU3.pdf}
            \end{minipage}
            \label{fig:skewiou3}
        }
    \end{center}
    \caption{The curves represent the relationship between SkewIoU and angle deviation $\Delta \theta$ under different aspect ratios. $k$ indicates the aspect ratio.}
    \label{fig:skewiou}
\end{figure*}

\subsection{DETR and Its Variants} 
DETR \cite{carion2020end} proposed a Transformer-based end-to-end object detector without using hand-designed components like prior anchor design and NMS. In recent years, DETR has progressed a lot and also exhibits its strong ability in object detection compared with classic detection methods \cite{zhu2021deformable, meng2021conditional, li2022dn, zhang2023dino, wang2022anchor, liu2022dab}. Deformable DETR \cite{zhu2021deformable} proposes a deformable attention module to sample the value of adaptive positions around the reference point and utilizes the multi-level features to mitigate the slow convergence and high complexity issues of DETR. DAB-DETR \cite{liu2022dab} provides explicit positional priors for each query to let the cross-attention module focus on a local region corresponding to a target object by using anchor box size.
DN-DETR \cite{li2022dn} and DINO \cite{zhang2023dino} design a denoising auxiliary task that bypasses the bipartite graph matching. This not only accelerates training convergence but also leads to a better training result.
In addition, there have been some DETR-based oriented object detectors \cite{ma2021oriented, dai2022ao2}. O$^2$DETR \cite{ma2021oriented} is the first attempt to apply DETR to the oriented object detection task and AO2-DETR \cite{dai2022ao2} introduces oriented proposal generation and refinement module into the transformer architecture to refine the features. 
Nevertheless, both of them predict angles using a simple regression way and do not address boundary discontinuity or embed angle information into DETR.

\begin{figure}[!tb]
    \begin{center}
        \subfloat[situation 1]{
            \begin{minipage}[t]{0.30\linewidth}
                \centering
                \includegraphics[width=1.0\linewidth]{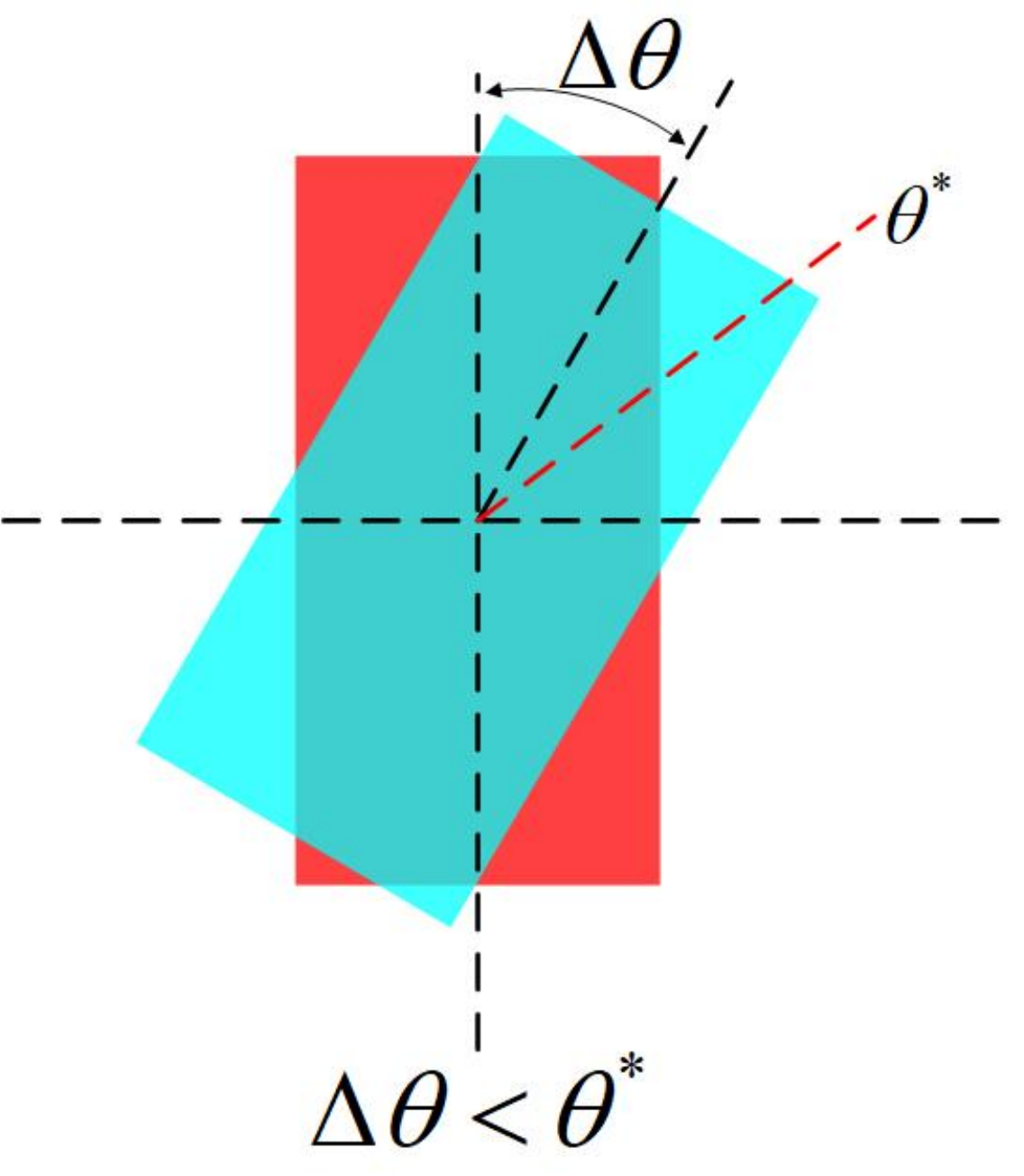}
            \end{minipage}%
        }
        \subfloat[boundary condition]{
            \begin{minipage}[t]{0.30\linewidth}
                \centering
                \includegraphics[width=1.0\linewidth]{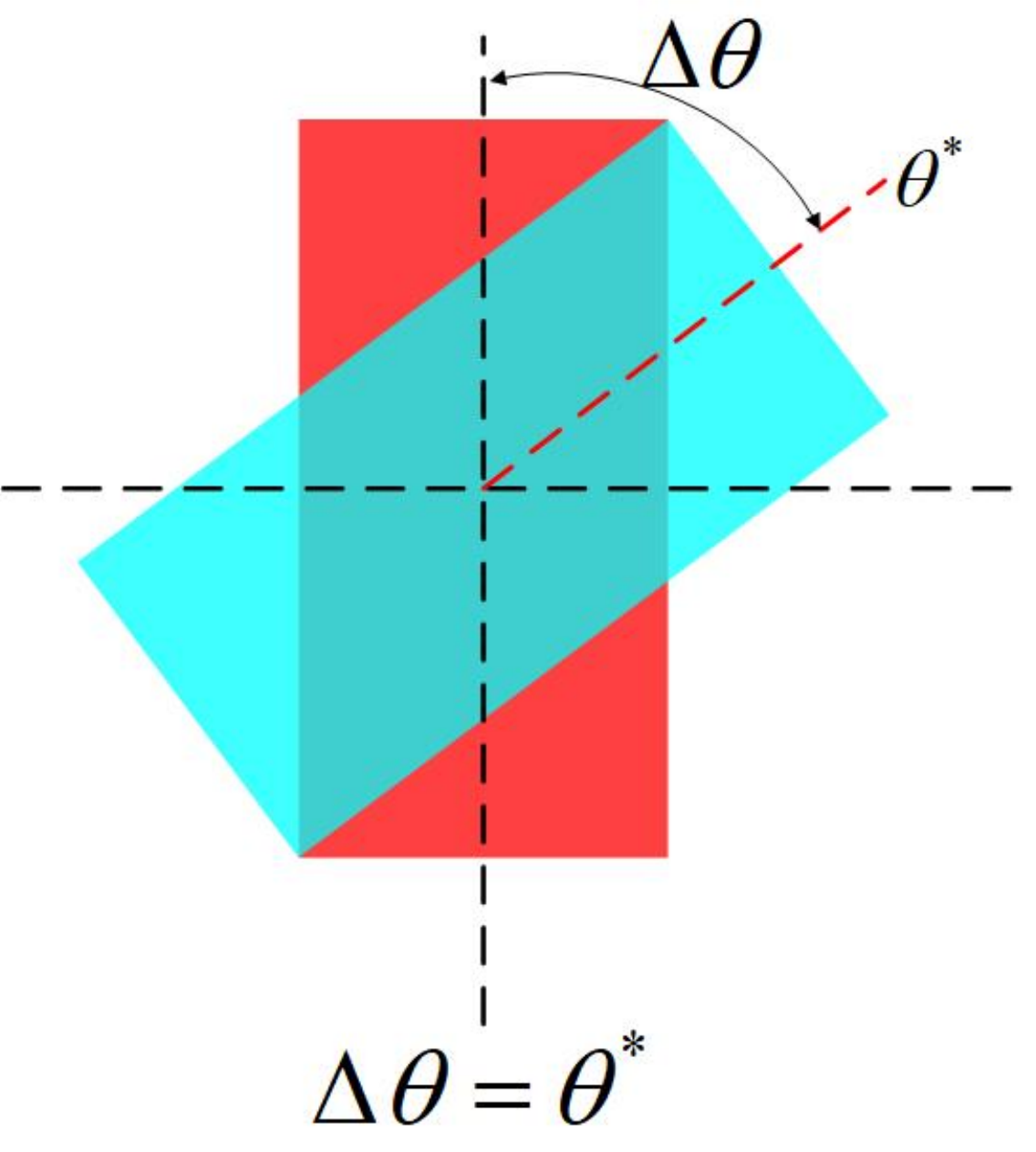}
            \end{minipage}%
        }
        \subfloat[situation 2]{
            \begin{minipage}[t]{0.30\linewidth}
                \centering
                \includegraphics[width=1.0\linewidth]{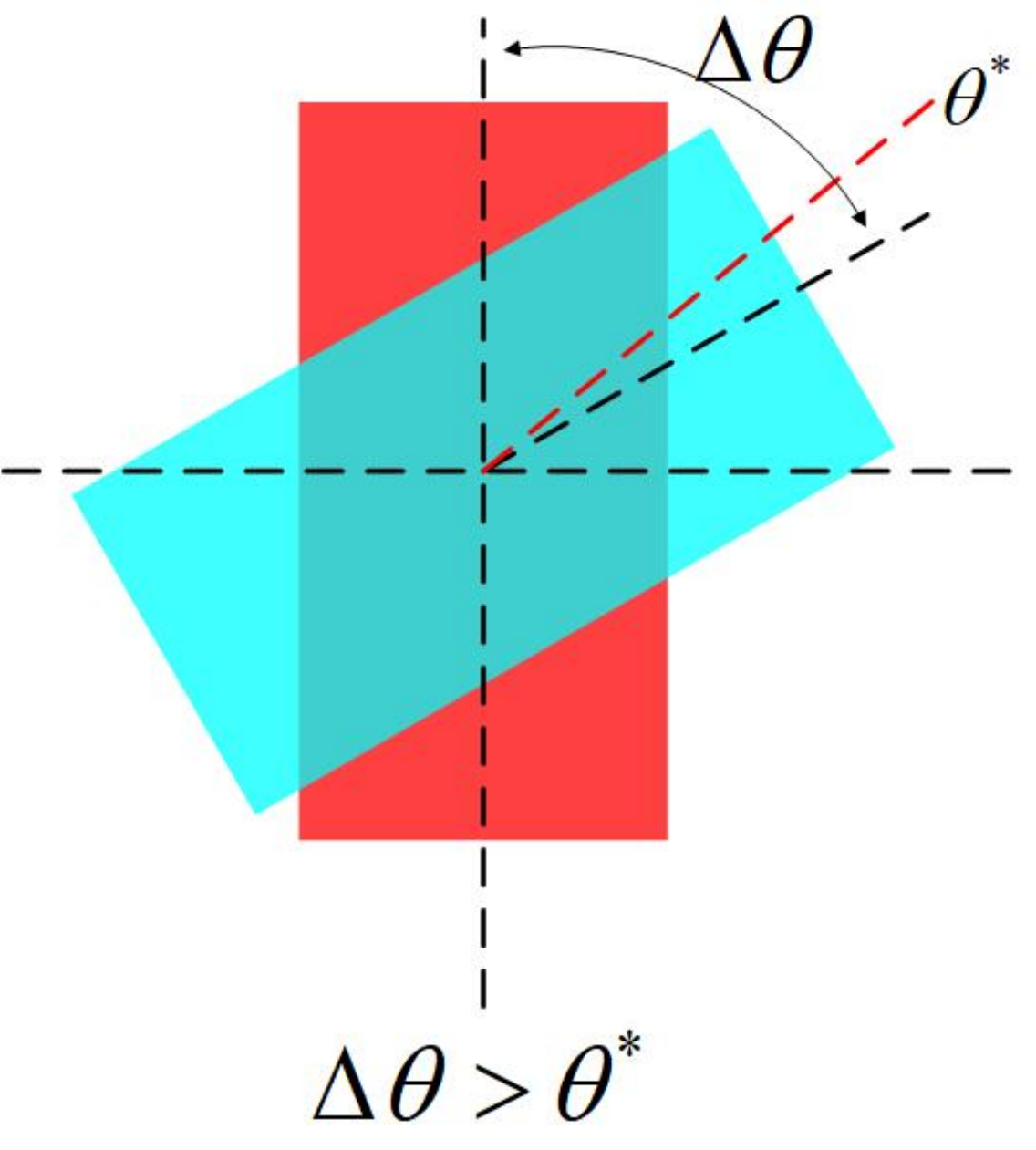}
            \end{minipage}%
        }
    \caption{Two situations for SkewIoU calculation. (a) The situation where $\Delta \theta < \theta^{*}$; (b) The boundary condition between situation 1 and situation 2; (c) The situation where $\Delta \theta > \theta^{*}$.}
    \label{fig:boundary_of_skewIoU}
    \end{center}
\end{figure}

\section{Rethinking on Oriented Object Detection}
In this section, we analyze the relationship between angle and aspect ratio. Additionally, we also analyze the shortcomings of currently used metric AP$_{50}$ in oriented object detection and emphasize the importance of high-precision oriented object detection.

\subsection{Angle and Aspect Ratio}
Objects with different aspect ratios have different sensitivities to angles.
In order to better observe the relationship between the angle and aspect ratio, we assume that there are two bounding boxes with the same center, width and height, and give the SkewIoU of these two boxes with different aspect ratios under different angle deviations, as shown in Fig.~\ref{fig:skewiou}. The SkewIoU can be calculated by Eq.~\ref{eq:skewiou}:

\begin{equation}
\begin{aligned}
    \text{SkewIoU}(k,\Delta \theta) = & 
    \left\{ \begin{array}{rcl}
    \frac{4k\tan \Delta \theta -m -n}{4k\tan \Delta \theta +m+n} & \Delta \theta \le 2\arctan \frac{1}{k} \\ 
    \frac{1}{2k\sin \Delta \theta -1} & \Delta \theta > 2\arctan \frac{1}{k}
    \end{array}\right. ,\\
    \small m=&(1-k\tan \frac{\Delta \theta }{2} )^{2} \tan^{2}  \Delta \theta ,\\
    \small n=&(\frac{-2\sin ^{2} \frac{\Delta \theta }{2}+k\sin \Delta \theta }{\cos \Delta \theta } )^{2} ,
\end{aligned}
\label{eq:skewiou}
\end{equation}
where $k \geq 1$ is the aspect ratio, 
and $\Delta \theta \in [0^\circ, 90^\circ]$ represents angle deviation, indicating the absolute value of the angle difference between two boxes. 
There is a critical angle boundary threshold ($\theta^{*}=2\arctan \frac{1}{k}$) as shown in Fig.~\ref{fig:boundary_of_skewIoU}.
The symmetrical nature depicted in Fig.~\ref{fig:skewiou} demonstrates that variations in $\Delta \theta$ exhibit bidirectional characteristics.

Fig.~\subref*{fig:skewiou1} shows the curves between SkewIoU and angle deviation under different aspect ratios. It can be seen that the SkewIoU variation trends of bounding boxes with different aspect ratios are obviously divided into two types according to metric AP$_{50}$ (if the SkewIoU between the prediction box and ground truth is greater than 0.5, then it will be judged as true positive), and the dividing boundary is $k=1.5$, as shown in Fig.~\subref*{fig:skewiou2} ($1 \leq k \leq 1.5$) and Fig.~\subref*{fig:skewiou3} ($k > 1.5$), respectively.
Specifically, Fig. \subref*{fig:skewiou2} shows that when the aspect ratio is smaller than 1.5, SkewIoU is always greater than 0.5 regardless of the angle deviation (see 
pink dashed line).
In contrast, when the aspect ratio is greater than 1.5, as shown in Fig.~\subref*{fig:skewiou3}, SkewIoU will decay rapidly with the increase of angle deviation, but the valid angle deviation still retains a wide range. 
In summary, objects with a small aspect ratio are less sensitive to angle deviation, whereas objects with a large aspect ratio are more sensitive but still exhibit a significant tolerance for angle deviation under AP$_{50}$.

\subsection{High-Precision Oriented Object Detection}

Considering that angle is a very important parameter in oriented object detection, the accuracy of its estimation will greatly affect the subsequent related tasks, such as object fine-grained recognition \cite{sun2022fair1m}, object heading estimation \cite{yang2018position,yang2022on}, etc., AP$_{50}$ seems not accurate enough for reflecting the performance of high-precision oriented object detection. Therefore, we advocate to use more stringent metric, e.g. AP$_{75}$\footnote{\textbf{Note:} The difficulty of achieving high AP$_{75}$ in oriented object detection is more difficult than that in horizontal object detection, because the rotated bounding box is more accurate with less redundant areas, thus more sensitive to errors.}, which is usually used in generic detection, to measure this challenging task. 
Under the AP$_{75}$ metric, not only are the prediction boxes required to be closer to the ground truth boxes, but the angle prediction requirement is also more stringent. 
As shown the 
gray dashed line
in Fig. \ref{fig:skewiou}, when AP$_{75}$ is adopted, regardless of the aspect ratio, 
the angle deviation should be controlled within a specific range; otherwise, they will not be judged as positive. The larger the aspect ratio, the narrower the range.

Tab. \ref{tab:ap50_vs_ap75} compares the accuracy of some oriented object detectors using AP$_{50}$ and AP$_{75}$, respectively. It can be seen that all detectors achieve a high performance in terms of AP$_{50}$ and the gap among them is small. However, the situation becomes different when AP$_{75}$ is used, some detectors may not good as other detectors whose performance on AP$_{50}$ are lower than them, e.g. S$^2$A-Net vs. Rotated ATSS. 
Therefore, AP$_{75}$ can further represent the performance of high-precision oriented object detection.

\begin{table}[!tb]
    \caption{Accuracy of some oriented object detectors on DOTA-v1.0 and DIOR-R datasets.}
    \begin{center}
        \resizebox{0.48\textwidth}{!}{
            \begin{tabular}{l|ccc|ccc} 
                \hline\hline
                \multirow{2}{*}{Method} & \multicolumn{3}{c|}{DOTA-v1.0} & \multicolumn{3}{c}{DIOR-R}  \\ 
                \cline{2-7}
                & \multicolumn{1}{c|}{AP$_{50}$} & \multicolumn{1}{c|}{AP$_{75}$} & \multicolumn{1}{c|}{AP$_{50:95}$} & \multicolumn{1}{c|}{AP$_{50}$} & \multicolumn{1}{c|}{AP$_{75}$} & \multicolumn{1}{c}{AP$_{50:95}$}\\
                \hline
                Rotated FCOS~\cite{tian2019fcos} & 72.45 & 39.84 & 41.02 & 62.00 & 36.10 & 37.61 \\
                S$^2$A-Net~\cite{han2021align} & 75.29 & 40.08 & 42.00 & 64.50 & 38.24 & 38.02\\
                Rotated Faster RCNN~\cite{ren2015faster} & 73.96 & 43.44 & 42.93 & 63.41 & 41.80 & 39.72\\
                KLD~\cite{yang2021learning} & 73.46 & 44.74 & 43.70 & 64.63 & 41.60 & 40.34 \\
                Rotated ATSS~\cite{zhang2020bridging} & 73.37 & 44.95 & 43.53 & 63.52 & 42.61 & 40.72\\
                GWD~\cite{yang2021rethinking} & 73.25 & 45.21 & 44.04 & 60.31 & 40.90 & 39.70 \\
                Oriented Reppoints~\cite{li2022oriented} & 74.38 & 46.56 & 44.57 & 66.31 & 44.36 & 42.81 \\
                \hline\hline
            \end{tabular}
        }
    \end{center}
    \label{tab:ap50_vs_ap75}
\end{table}

As for the AP$_{50:95}$, which is also often adopted in generic detection, it is no doubt a stricter and more comprehensive metric, but we believe that AP$_{75}$ could reflect high-precision oriented object detection more directly. AP$_{50:95}$ contains a large number of metrics. Under the metrics AP$_{50}$ or AP$_{55}$, when the angle prediction is not accurate, as shown in Fig.~\ref{fig:inaccurate_angle_prediction}, a relatively high value could still be reached. Conversely, under the metrics AP$_{90}$ or AP$_{95}$, the performance of oriented detectors degenerates a lot, which is of no significance for comparison in current research. Therefore, AP$_{75}$, as an intermediate metric, is more balanced and suitable for high-precision oriented object detection. Meanwhile, there is an average operation in AP$_{50:95}$, which makes metrics like AP$_{50}$ and AP$_{55}$ have a great influence on the overall value.

In summary, it is necessary and meaningful to pay more attention to high-precision oriented object detection.

\section{Method}\label{sec:method}

\begin{figure*}[!tb]
    \begin{minipage}[t]{1.0\linewidth}
        \centering
        \includegraphics[width=1.0\linewidth]{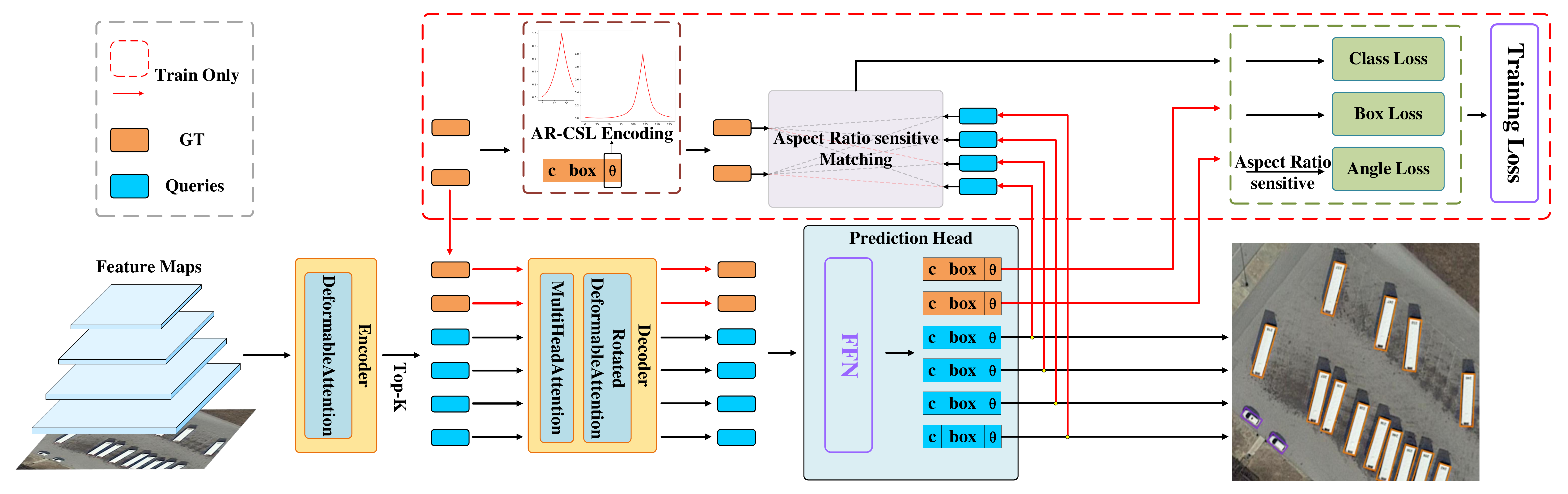}
    \end{minipage}%
    \label{fig:struct}
\centering
    \caption{The framework of the proposed ARS-DETR. ‘GT’ means ground truth. ‘Train Only’ means it only works during the training process and will be removed during the inference.
    \label{fig:structure}}
\end{figure*}

In this section we mainly design our methods around the relationship between the angle and aspect ratio. 
We propose a new angle classification method, AR-CSL, to dynamically adjust the smoothing process according to the aspect ratio. Then we adopt the Deformable DETR\cite{zhu2021deformable} as the detection architecture and develop it with rotated deformable attention module, denoising training strategy, aspect ratio sensitive matching and loss to adapt to the oriented object detection.

\subsection{Overview}

Fig.~\ref{fig:structure} shows the framework of the proposed ARS-DETR. Given an image, a backbone is firstly used to extract feature maps. The backbone will generate hierarchical feature maps and the last three stages of outputs are used. Then, the 1$\times$1 convolution is adopted to map their channels to the uniform dimension. Additionally, the lowest resolution feature map is obtained via a 3$\times$3 convolution on the final feature map. 
Then the multi-scale feature maps, embed with the 2D positional encoding, are fed into Encoder. 
Without the use of top-down structure in FPN\cite{lin2017feature}, multi-scale Deformable Attention can exchange the information among multi-scale feature maps and further refine these feature maps.
The output of Encoder will then generate a large number of proposals and be used in Decoder. Next, the Top-K scoring proposals are picked as object queries and transformed into output embeddings by MultiHead Attention and multi-scale Rotated Deformable Attention in Decoder. Finally, the prediction head further decode the output embeddings from decoder into class labels, angle labels and horizontal box coordinates. Additionally, during the training process, we utilize the ground truth with noise as queries to participate in the training to stabilize the training. 
At the same time, we adopt Aspect Ratio sensitive Matching (ARM) to adjust the influence of angle in matching process and adopt Aspect Ratio sensitive Loss (ARL) to adjust the training strategy for different aspect ratio objects when calculating angle loss.

\subsection{Aspect Ratio Aware Circle Smooth Label}
\subsubsection{Rethinking on Circular Smooth Label}
Instead of using regression-based loss function, Circular Smooth Label (CSL) \cite{yang2020arbitrary} transforms angle prediction into a classification task so that the boundary problem naturally disappears. 
As shown in Fig.~\subref*{fig:csl_arsl1} and Fig.~\subref*{fig:csl_arsl2}, CSL divides the angle into 180 categories and treats the first angle category and the last angle category as adjacent angle categories to eliminate the impact of boundary discontinuity. Then, it adopts Gaussian window function to smooth the angle category label of the objects so as to reflect the correlation among adjacent angle categories and make it have a certain tolerance for angle estimation error. The expression of CSL is as follows:

\begin{equation}
    \begin{aligned}
    CSL(t) & = \left\{\begin{matrix}
    g(t), & \theta -r<t<\theta +r\\
    0, & otherwise 
    \end{matrix}\right. ,
    \end{aligned}
    \label{eq:csl}
\end{equation}
where $g()$ is window function, $t$ is the angle represented by label, $r$ is the radius of the window function, and $\theta$ is the angle of ground truth.

Although CSL has made some progress, it still has two drawbacks which will behind its performance:
\begin{itemize}
    \item \textbf{Fixed label function.} CSL adopts a fixed radius Gaussian function to learn the correlation among adjacent angles and smooth the label, without considering objects' aspect ratio, as shown in Fig.~\subref*{fig:csl_arsl1}. 
    However, it can be obviously seen from the Fig.~\ref{fig:skewiou} that the SkewIoU of objects with different aspect ratio differs a lot in adjacent angles. 
    Therefore, the correlation among adjacent angles should not be rigid and Gaussian window is likely not the best choice for all objects.
    \item \textbf{Angle discrete granularity insensitivity.} CSL is also insensitive to the angle discrete granularity. 
    When the angle discrete granularity $\omega$ is 1, indicating the angle is divided into 180 categories, the smoothing result is shown in Fig.~\subref*{fig:csl_arsl1}. In contrast, when the angle discrete granularity $\omega$ is 15, indicating the angle is divided into 12 categories, the smoothing result is shown in Fig.~\subref*{fig:csl_arsl5}. It can be seen from these two results that the smoothing outcomes of CSL remain consistent under different angle discrete granularities, which is clearly unreasonable.
    With the increase of $\omega$, the correlation among adjacent angles will become weaker, while CSL is insensitive to this and will still give the same smoothing values to the adjacent angle categories. Hence, the correlation among adjacent angles under different angle discrete granularity should also be taken into account.
    \item \textbf{Hyperparameter introduction.} The radius of window function will affect the final performance to some extent. As a hyperparameter, it is a thorny problem to determine the best value of the radius when the $\omega$ changes.
\end{itemize}

\subsubsection{Design of Aspect Ratio Aware Circle Smooth Label} 
\par
According to the above analysis, the fixed window function and hyperparameter (i.e. radius) hurt the applicability of classification-based oriented object detectors to some extent. 
In this subsection, we will address the aforementioned issues from the perspective of the encoding form.

Considering that SkewIoU can dynamically reflect the correlation among adjacent angles of different objects, we design an Aspect Ratio aware Circle Smooth Label (AR-CSL) technique to obtain a more reasonable angle prediction, using the SkewIoU instead of a fixed window function to smooth the label. Specifically, we calculate the SkewIoU of the bounding boxes under each angle deviation according to Eq.~\ref{eq:skewiou}, and take the calculated values as the label of the current angle category bin.

\begin{figure*}[!tb]
    \centering
    \subfloat[CSL encoding in all objects (flatly unfolded)]{
        \begin{minipage}[t]{0.45\linewidth}
            \centering
            \includegraphics[width=0.98\linewidth]{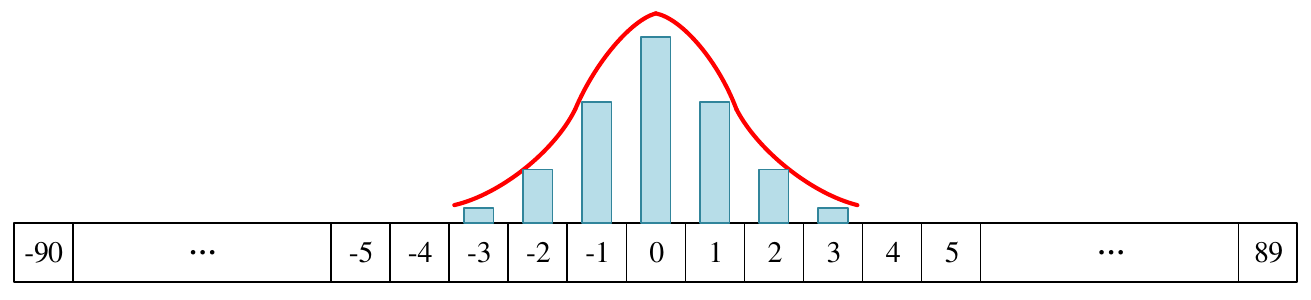}
        \end{minipage}%
        \label{fig:csl_arsl1}
    }
    \subfloat[Circular smooth label]{
        \begin{minipage}[t]{0.45\linewidth}
            \centering
            \includegraphics[width=0.30\linewidth]{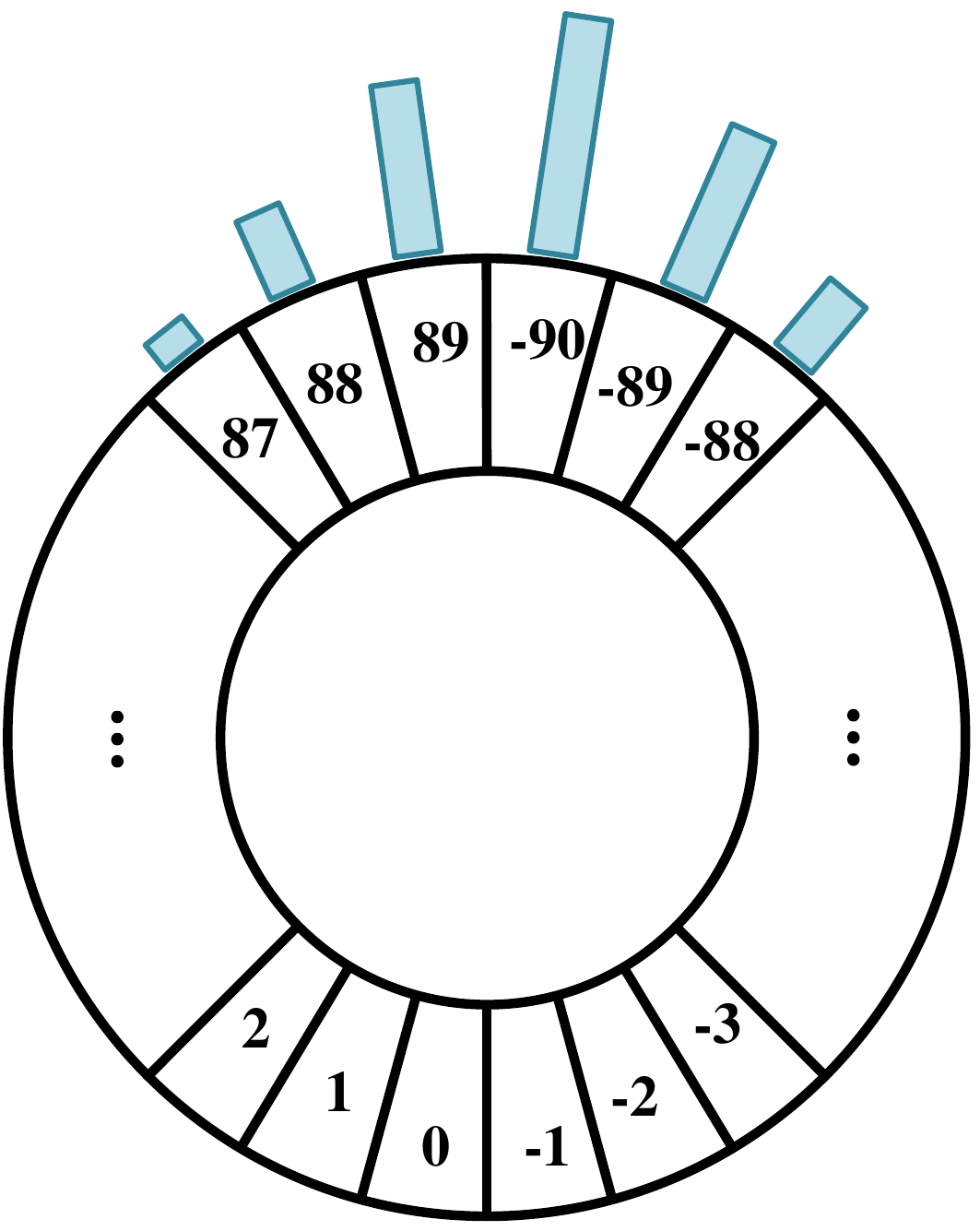}
        \end{minipage}%
        \label{fig:csl_arsl2}
    }\\
    \subfloat[AR-CSL encoding in small aspect ratio objects]{
        \begin{minipage}[t]{0.45\linewidth}
            \centering
            \includegraphics[width=0.98\linewidth]{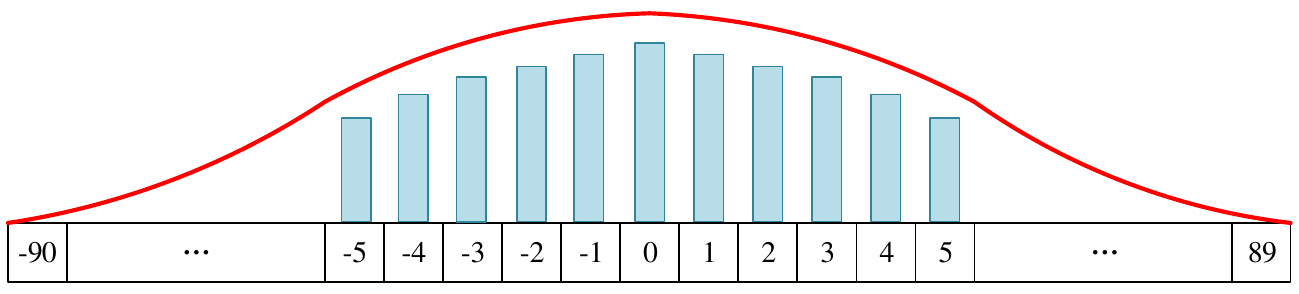}
        \end{minipage}
        \label{fig:csl_arsl3}
    }
    \subfloat[AR-CSL encoding in large aspect ratio objects]{
        \begin{minipage}[t]{0.45\linewidth}
            \centering
            \includegraphics[width=0.98\linewidth]{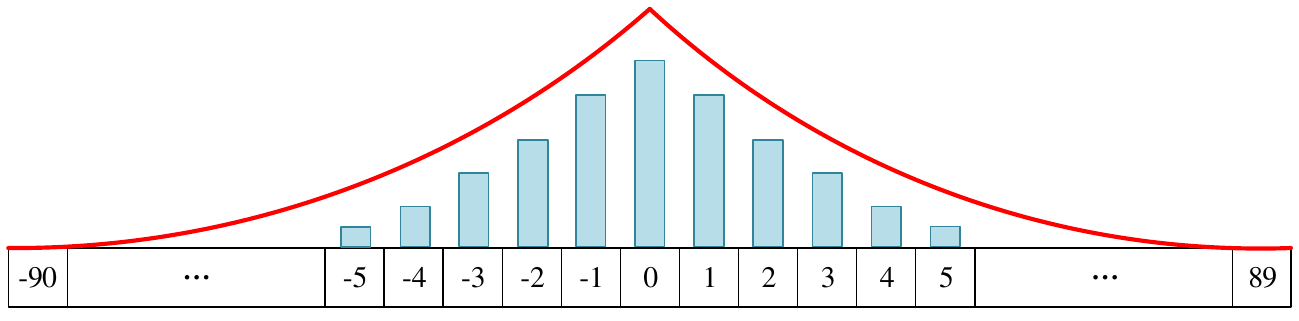}
        \end{minipage}
        \label{fig:csl_arsl4}
    }\\
    \subfloat[CSL encoding in large angle discrete granularity]{
        \begin{minipage}[t]{0.45\linewidth}
            \centering
            \includegraphics[width=0.98\linewidth]{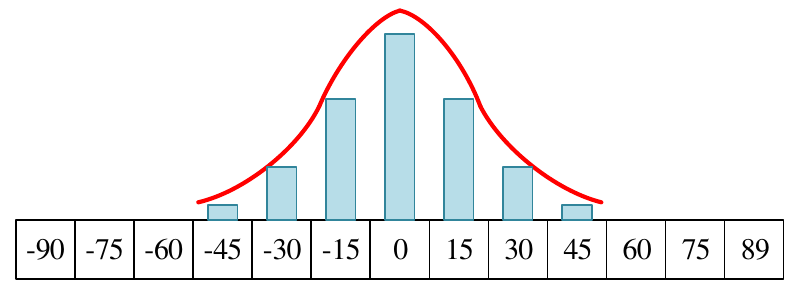}
        \end{minipage}
        \label{fig:csl_arsl5}
    }
    \subfloat[AR-CSL encoding in large angle discrete granularity]{
        \begin{minipage}[t]{0.45\linewidth}
            \centering
            \includegraphics[width=0.98\linewidth]{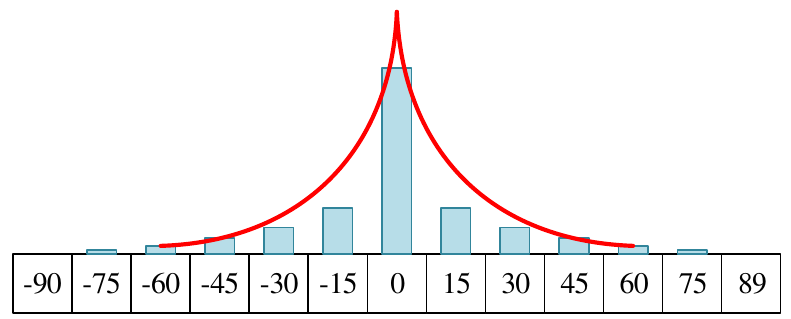}
        \end{minipage}
        \label{fig:csl_arsl6}
    }
\centering
    \caption{The comparison of two encoding methods in objects with different aspect ratio at each angle deviation. For the convenience of comparison, the labels in the Fig. (a) and Fig. (c)-(f) are flatly unfolded, otherwise they should be circular like (b).
    (a) For CSL, a Gaussian window with a fixed window radius will be adopted to smooth the angle label, regardless of the objects' aspect ratio. (c)-(d) For AR-CSL, objects with different aspect ratio will be considered and it will use a more reasonable smoothing strategy to reflect the correlation among the adjacent angles. (e) For CSL, angle discrete granularity $\omega$ will be overlooked and will give the same smoothing values under different angle discrete granularity $\omega$ (d) For AR-CSL, smoothing values are calculated dynamically according to the angle deviation and will vary under different angle discrete granularity $\omega$.}
    \vspace{-8pt}
    \label{fig:csl_vs_arsl}
\end{figure*}

\begin{figure}[tb]
    \centering
    \subfloat[Simple method]{
        \begin{minipage}[t]{0.47\linewidth}
            \centering
            \includegraphics[width=0.98\linewidth]{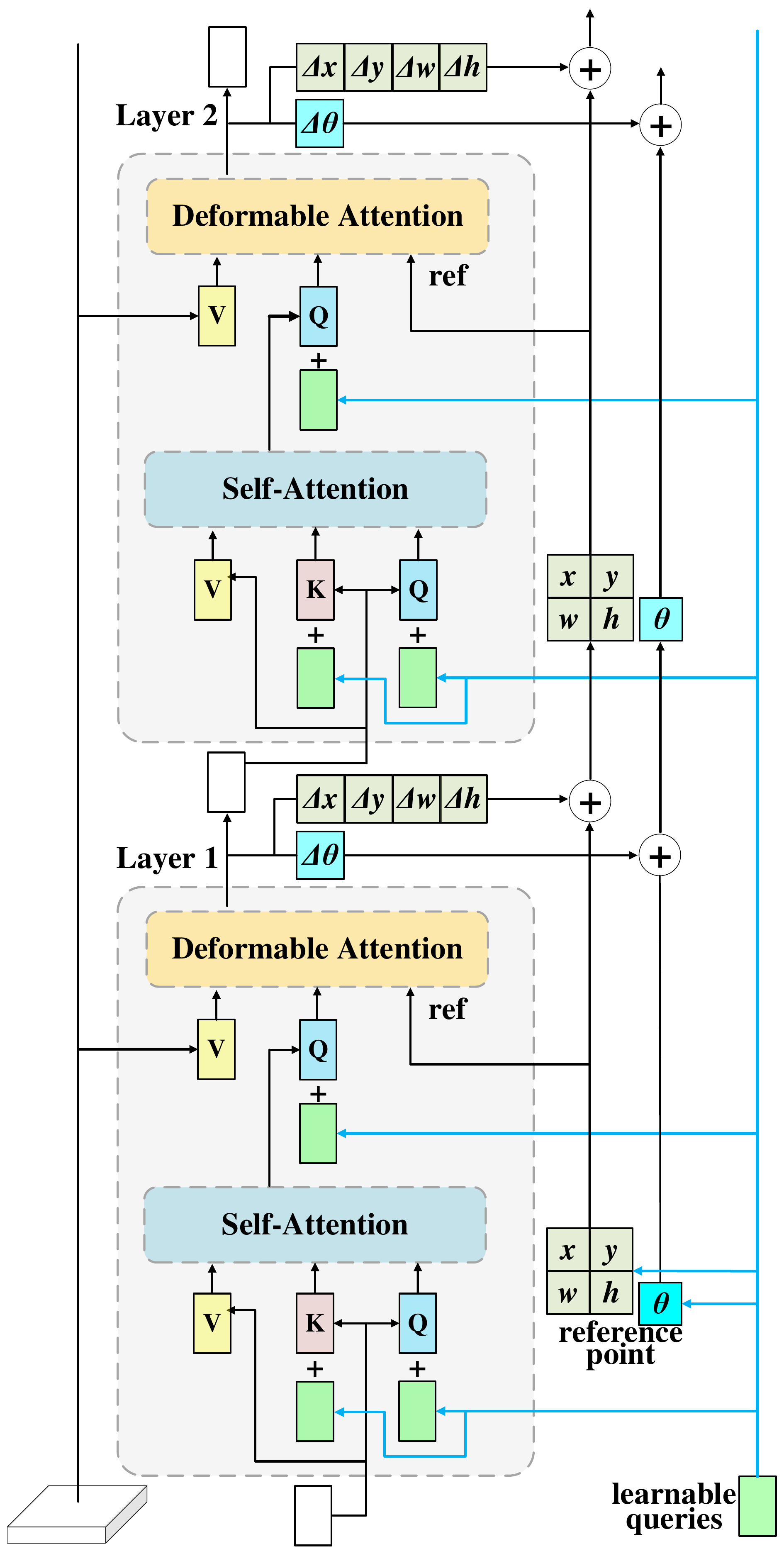}
        \end{minipage}%
        \label{fig:architecture1}
    }
    \subfloat[Ours]{
        \begin{minipage}[t]{0.47\linewidth}
            \centering
            \includegraphics[width=0.98\linewidth]{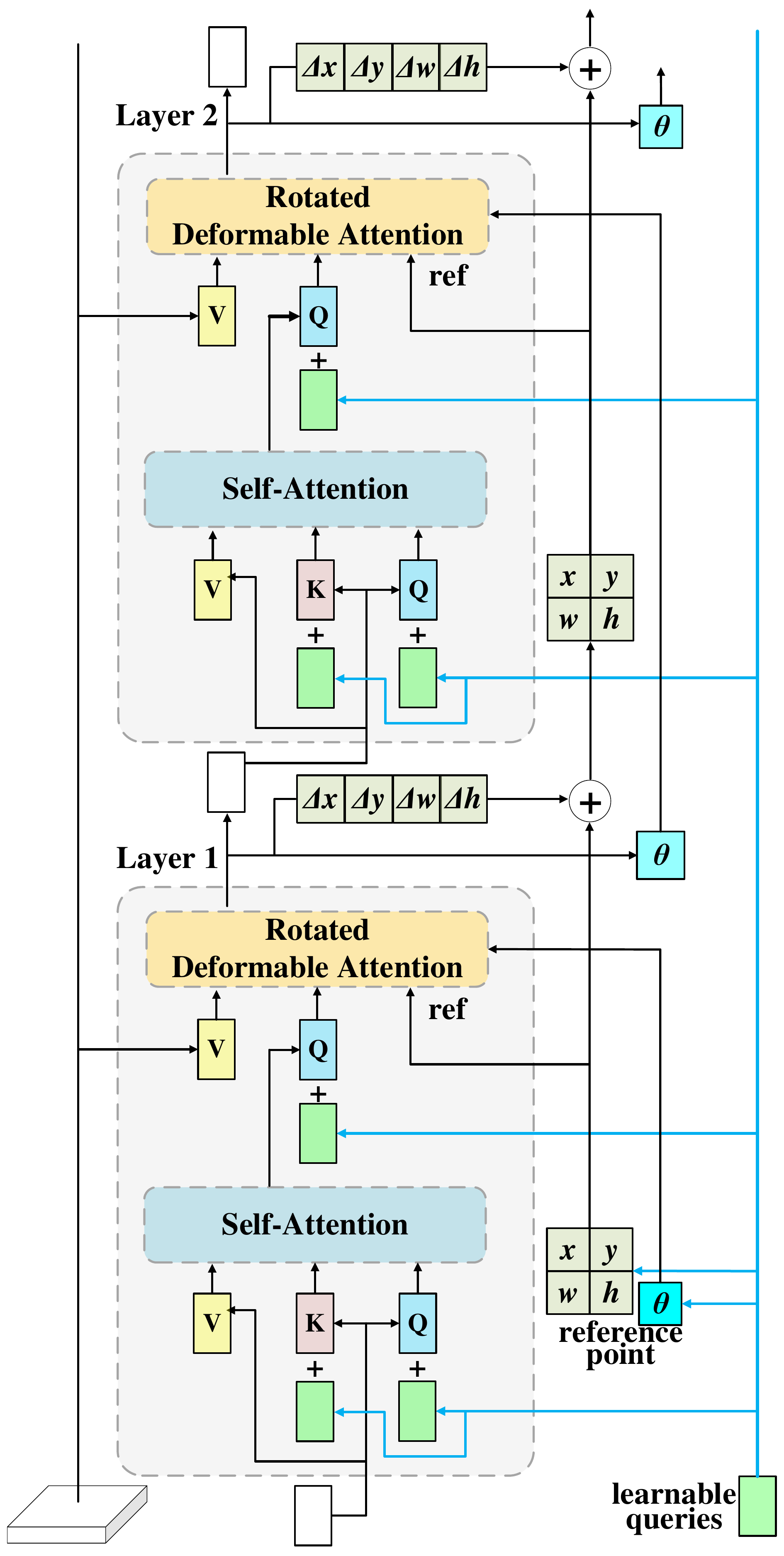}
        \end{minipage}%
        \label{fig:architecture2}
    }
    \centering
    \caption{Two methods to iterate the angle information in the DETR. (a) In the simple way, although the angle information is updated iteratively after each layer, it is not embed into DETR. (b) In our proposed way, the angle information will be replaced with a new value after each layer and the angle information will assist in aligning features.}
    \label{fig:architecture}
\end{figure}

Then, we normalize the SkewIoU values by maximum and minimum normalization method, shown as follows:
\begin{equation}
    \begin{aligned}
        \begin{matrix}
        AR-CSL(k, t)=\frac{SkewIoU(k, \Delta \theta ) - SkewIoU(k)_{min} }{1-SkewIoU(k)_{min}} ,\\
        \Delta \theta = \left | t-\theta  \right | ,
        \end{matrix}
    \end{aligned}
    \label{eq:arsl}
\end{equation}
where $k$ is aspect ratio of ground truth, $t$ is the angle represented by label, $\theta$ is the angle of ground truth and $\Delta \theta \in [0^\circ, 90^\circ]$ is the angle deviation.

Compared with CSL, the proposed AR-CSL has the following advantages:
\begin{itemize}
    \item \textbf{Dynamic label function.} 
    The smoothing values are dynamically calculated according to the aspect ratios of objects by using SkewIoU, as shown in Fig.~\subref*{fig:csl_arsl3}-\subref*{fig:csl_arsl4}.
    \item \textbf{Angle discrete granularity sensitivity.} 
    Because the angle deviation of different angle categories will be accounted for according to the Eq.~\ref{eq:arsl}, the smoothing values in adjacent categories will vary with alterations in angle discrete granularity, as shown in Fig.~\subref*{fig:csl_arsl4}-\subref*{fig:csl_arsl6}.
    \item \textbf{Hyperparameter free.} According to Eq.~\ref{eq:skewiou} and Eq.~\ref{eq:arsl}, no hyperparameters are introduced. 
    This makes the proposed method more convenient to use.
\end{itemize}

\subsection{Rotated Deformable Attention Module}

Fig.~\subref*{fig:architecture1} shows a simple DETR-based oriented detector \cite{ma2021oriented, dai2022ao2}. This detector merely adds an additional angle parameter in the prediction head to accomplish rotated bounding box estimation. Nevertheless, it fails to embed the angle information into the detector to exploit the maximum potential of the detector, resulting in feature misalignment. To address this, we present a Rotated Deformable Attention module (RDA).

Given an input feature map $ x\in \mathbb{R} ^{C\times H\times W}$, let $ q\in \Omega _{q} $ index a query element with representation feature $ z_{q} \in \mathbb{R} ^{C} $ and reference box $ b_{q} = \left [ p_{q},w_{q},h_{q},\theta _{q} \right ] = \left [ (x_{q},y_{q} ),w_{q},h_{q},\theta _{q} \right ]$, where $ p_{q} =(x_{q}, y_{q} )\in \left [ 0,1 \right ] ^{2} $ is the centric point of the reference box and $ w_{q} \in \left [ 0,1 \right ] $, $ h_{q} \in \left [ 0,1 \right ] $, $ \theta _{q} \in \left [ -\frac{\pi }{4}, \frac{\pi }{4} \right ] $ are the width, height and angle of the reference box respectively. The rotated deformable attention feature is calculated as follows:
\begin{equation}
    \begin{aligned}
        RDA(z_{q}, p_{q}, x)=\sum_{m=1}^{M}W _{m} \left [\sum_{k=1}^{K}A_{mqk}\cdot W_{m}^{'}x(p_{q}+\Delta p_{mqk}  )  \right ] , 
    \end{aligned}
\end{equation}
where $m$ indexes the attention head, and $k$ indexes the sampled points. $M$ is the total head number and $K$ is the total sampled points number. We use $M=8$ and $K=4$ following \cite{zhu2021deformable}. $ W_{m}^{'} \in \mathbb{R} ^{\frac{C}{M}\times C} $ and $ W_{m} \in \mathbb{R} ^{C\times \frac{C}{M}} $ are learnable weights. $ \Delta p_{mqk} $ and $ A_{mqk} $ denote the sampling offset and attention weight of the $ k^{th} $ sampling point in the $ m^{th} $ attention head, respectively. The scalar attention weight $ A_{mqk} $  lies in the range $\left [ 0,1 \right ] $, normalized by $\sum_{k=1}^{K}A_{mqk}  =1$.  For each $ \Delta p_{mqk} $, it is calculated by:
\begin{equation}
    \begin{aligned}
        \Delta p_{mqk}=\frac{(w,h)}{2K}(z_{q}f_{mk}+r_{mk})R^{T} (\theta _{q}) ,
    \end{aligned}
\end{equation}
where $f_{mk}\in \mathbb{R}  ^{C\times 2} $ is the projection matrices. $r_{mk}$ is the bias of the $ k^{th} $ sampling point in the $ m^{th} $ attention head and is calculated by:
\begin{equation}
    \begin{aligned}
        r_{mk}=\frac{k(cos(\frac{2\pi m}{M} ), sin(\frac{2\pi m}{M}))}{max(\left | cos(\frac{2\pi m}{M} ) \right | , \left | sin(\frac{2\pi m}{M} ) \right | )} ,
    \end{aligned}
\end{equation}
$R(\theta _{q} )=(cos\theta _{q}, -sin\theta _{q};sin\theta _{q}, cos\theta _{q})^{T}$ is the rotation matrix.

In this way, we can obtain the dynamic sampling points $p_{q} +\Delta p_{mqk} $ and constrain them as much as possible within $b_{q} $ to extract aligned features.

\begin{figure}[!tb]
    \centering
    \subfloat[Deformable Attention]{
        \begin{minipage}[t]{0.45\linewidth}
            \centering
            \includegraphics[width=0.98\linewidth]{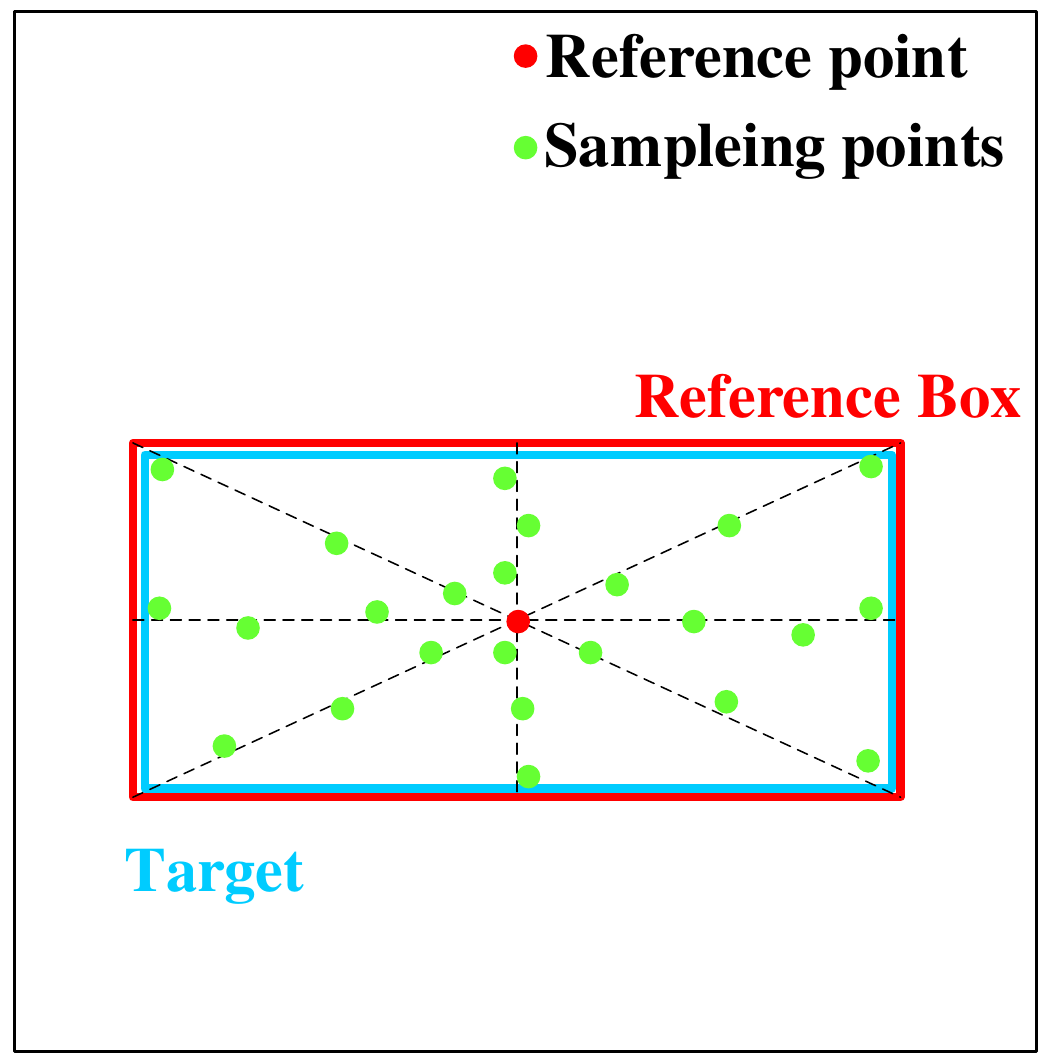}
        \end{minipage}%
        \label{fig:rda1}
    }
    \subfloat[Misalign]{
        \begin{minipage}[t]{0.45\linewidth}
            \centering
            \includegraphics[width=0.98\linewidth]{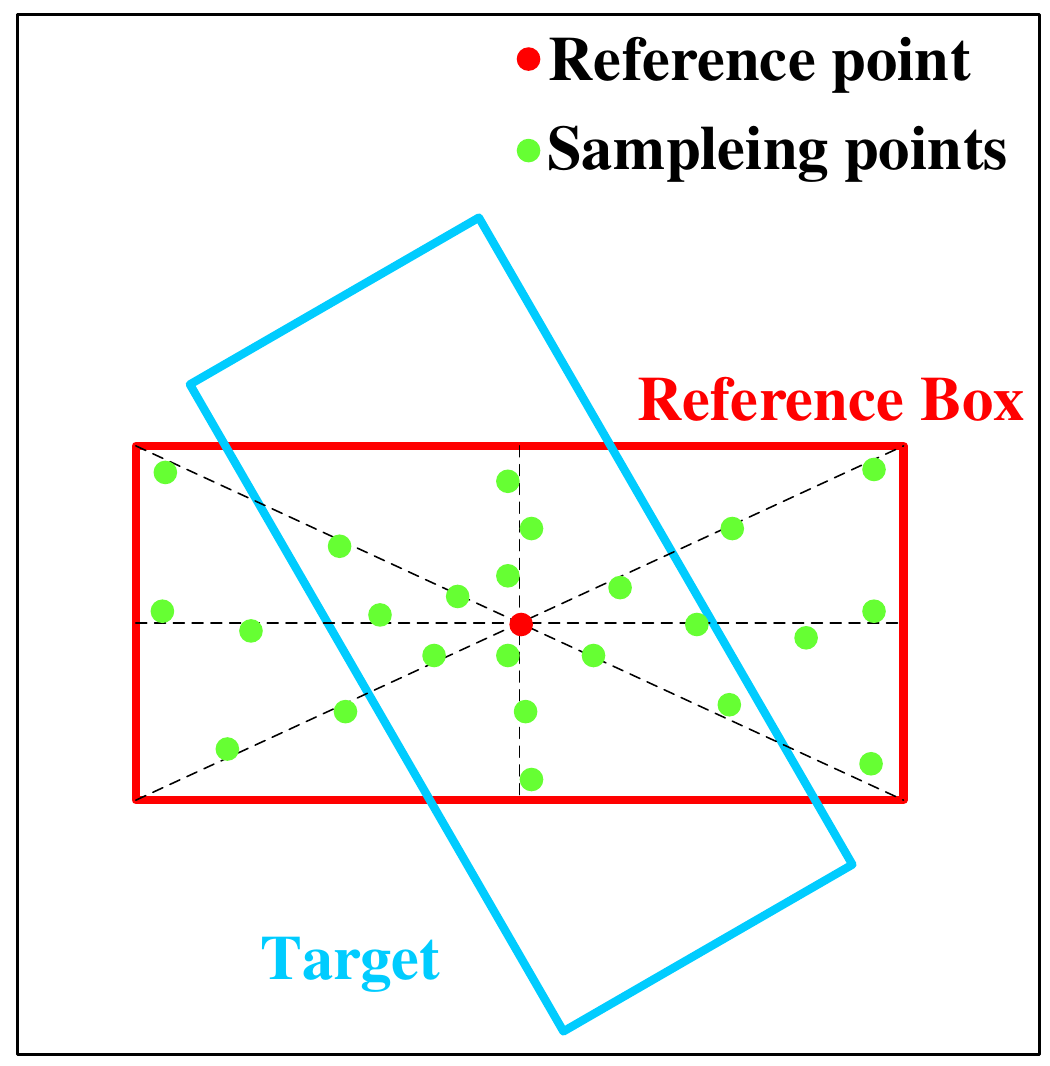}
        \end{minipage}%
        \label{fig:rda2}
    }\\
    \subfloat[Align]{
        \begin{minipage}[t]{0.45\linewidth}
            \centering
            \includegraphics[width=0.98\linewidth]{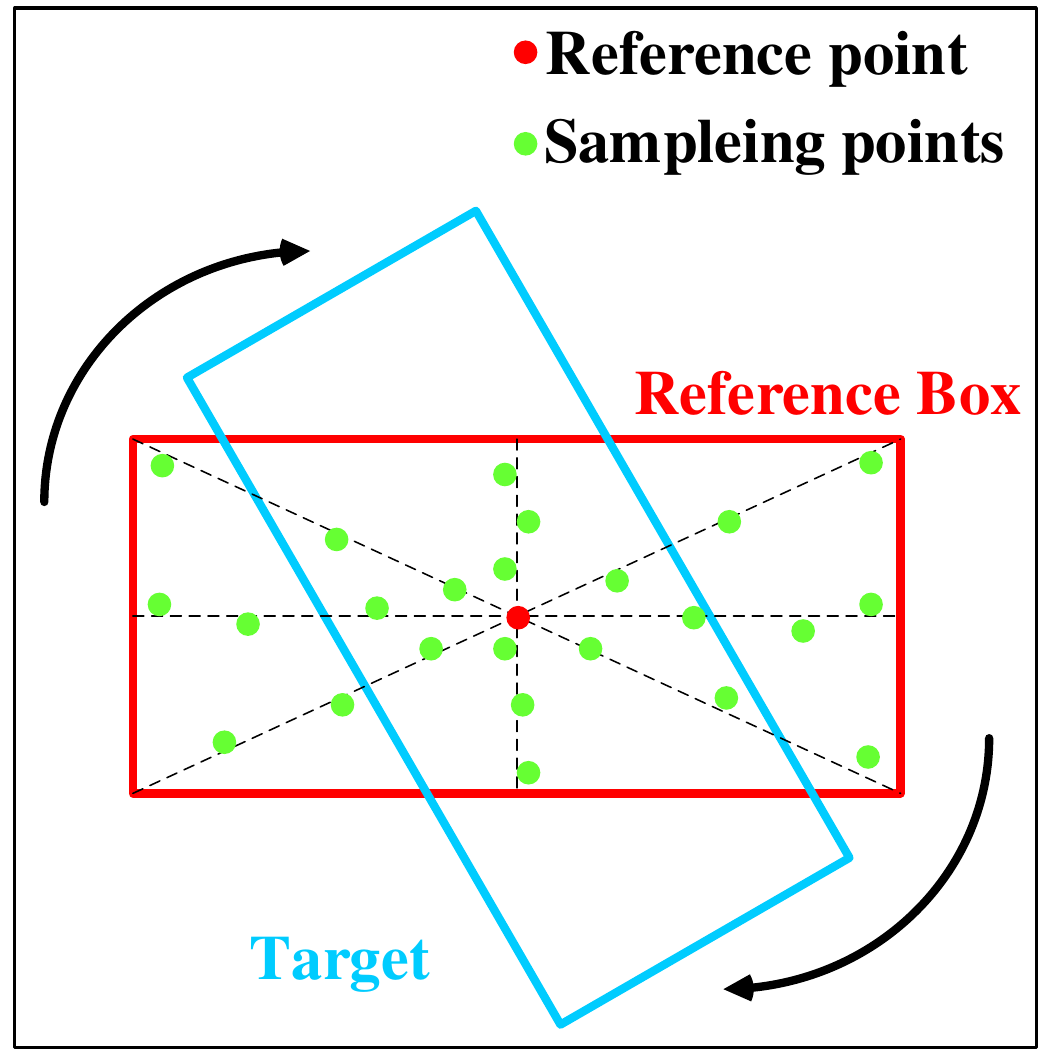}
        \end{minipage}
        \label{fig:rda3}
    }
    \subfloat[Rotated Deformable Attention]{
        \begin{minipage}[t]{0.45\linewidth}
            \centering
            \includegraphics[width=0.98\linewidth]{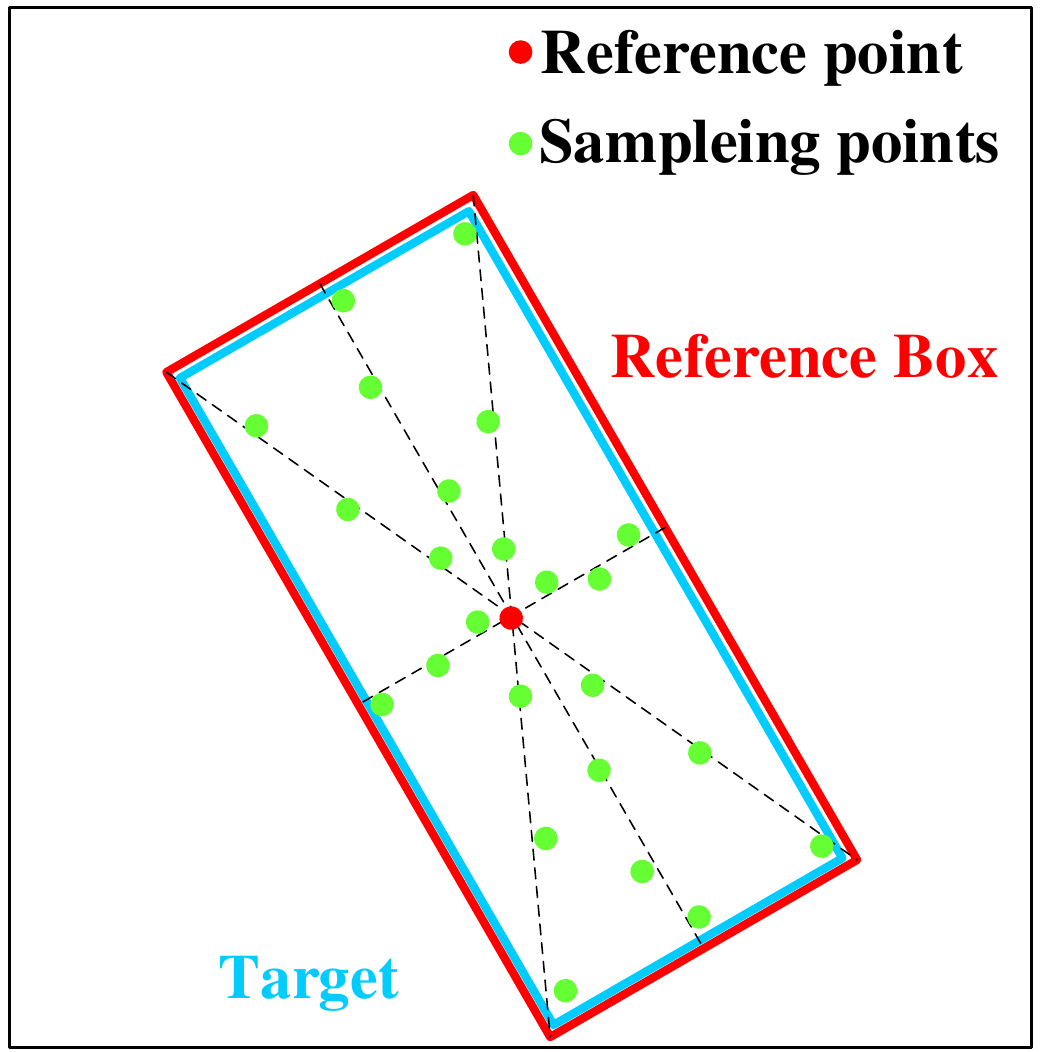}
        \end{minipage}
        \label{fig:rda4}
    }
\centering
    \caption{Illustration of the misalignment in Deformable Attention and the alignment in Rotated Deformable Attention.}
    \label{fig:misalign}
\end{figure}

\begin{figure}[!tb]
    \centering
    \subfloat[Deformable Offsets]{
        \begin{minipage}[t]{0.33\linewidth}
            \centering
            \includegraphics[width=0.98\linewidth]{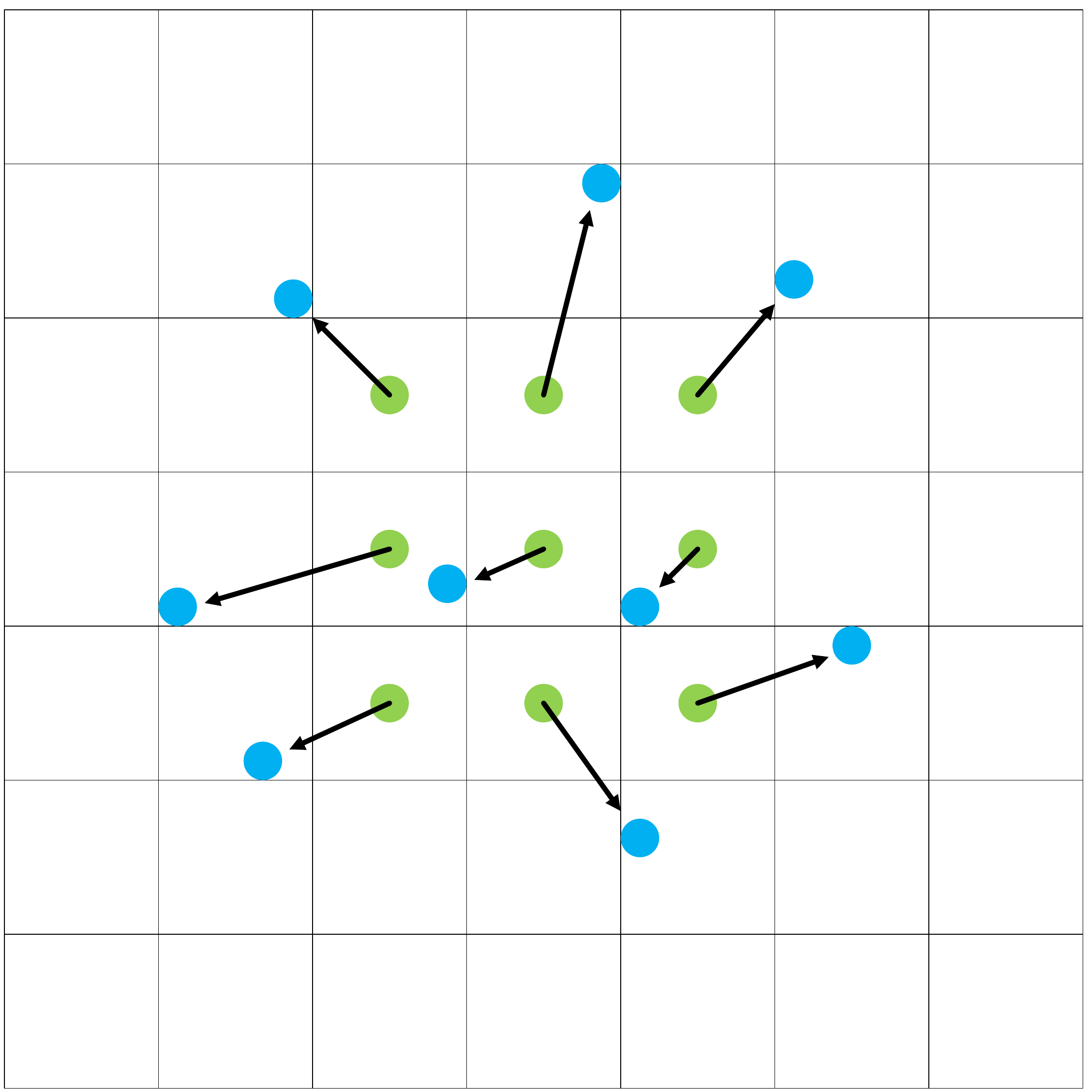}
        \end{minipage}%
        \label{fig:deformable_offsets}
    }
    \subfloat[Fixed Offsets]{
        \begin{minipage}[t]{0.33\linewidth}
            \centering
            \includegraphics[width=0.98\linewidth]{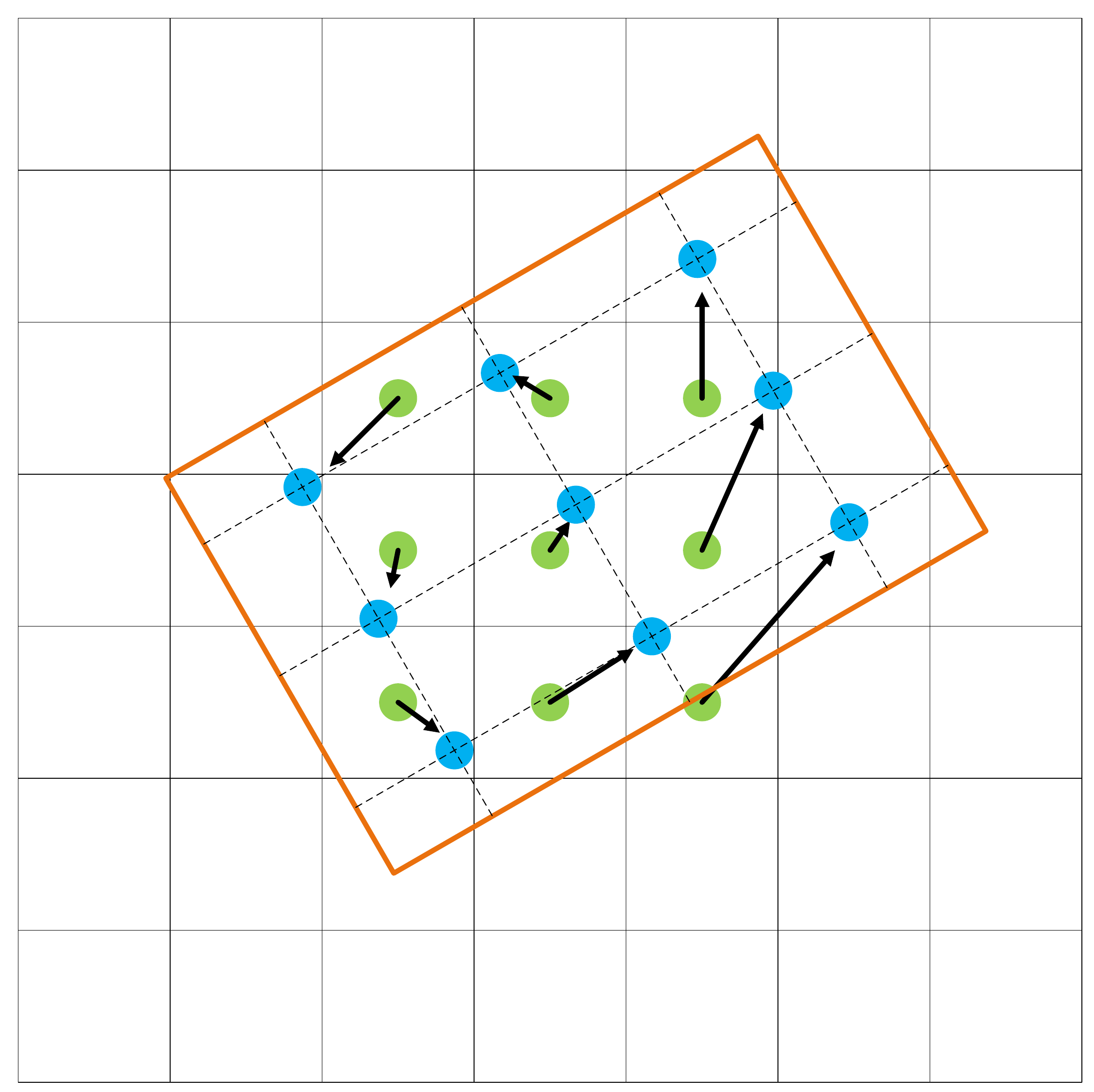}
        \end{minipage}%
        \label{fig:fixed_offsets}
    }
    \subfloat[RDA]{
        \begin{minipage}[t]{0.33\linewidth}
            \centering
            \includegraphics[width=0.98\linewidth]{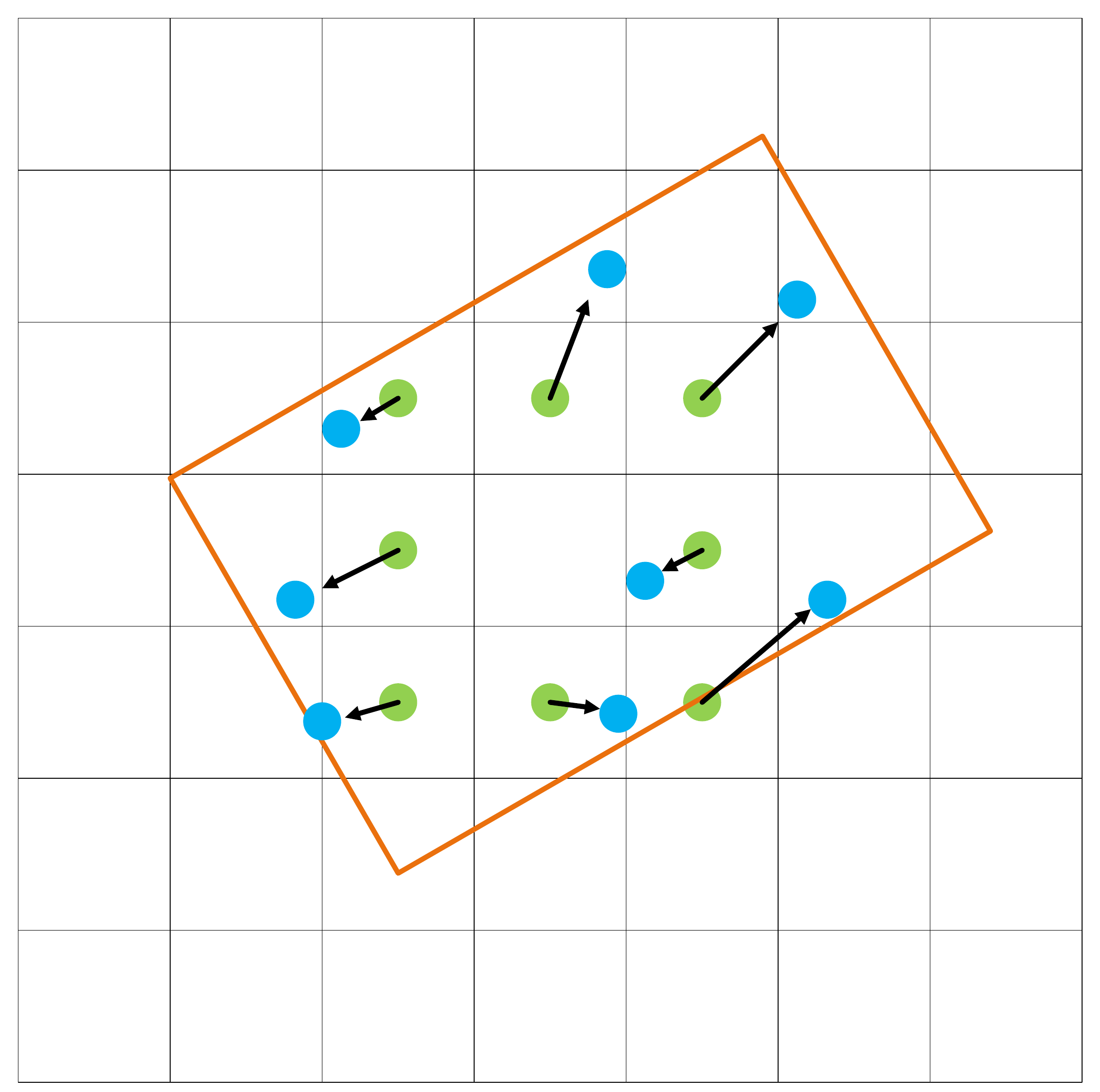}
        \end{minipage}
        \label{fig:cmp_RDA}
    }
\centering
    \caption{Illustration of different sampling methods. The orange rectangles mean bounding boxes. The black arrows mean the offset field. The green dots are regular sampling locations and the blue dots are sampling locations with offset. (a) deformable offsets; (b) fixed offsets with rotation in bounding box; (c) deformable offsets with rotation in bounding box.}
    \label{fig:comparison_of_RDA}
\end{figure}

As depicted in Fig.~\subref*{fig:rda1}, the sampling points in Deformable Attention Module will be adjusted according to the corresponding reference box, so that the sampling points will be restricted within the reference box and fall within the object as far as possible. However, as shown in Fig.~\subref*{fig:rda2}, 
when the object is of the oriented type, the sampling points cannot accurately align with the object if the horizontal reference box \cite{dai2022ao2} is still used.
Therefore, we design the Rotated Deformable Attention Module to align the sampling points with features by rotating the sampling points according to the embedded angle information, as shown in Fig.~\subref*{fig:rda3} and Fig.~\subref*{fig:rda4}. 
Moreover, instead of refining the angle layer by layer, we predict a new angle after each layer independently, as shown in Fig.~\subref*{fig:architecture2}.

As shown in Fig.~\ref{fig:comparison_of_RDA}, we compare two other sampling methods with our RDA. Fig.~\subref*{fig:deformable_offsets} shows the sampling method in \cite{dai2017deformable}. It learns deformable offsets to augment the spatial sampling locations, but it may sample from wrong locations with weak supervision, especially for densely packed objects. Fig.~\subref*{fig:fixed_offsets} shows the sampling methods in \cite{ding2019learning, han2021align}. It rotates the sampling points with the angle of the bounding box to align features, but the sampling positions are fixed. Our RDA also rotates the sampling points but it also learns deformable offsets and will constrain them according to the width and height of the bounding box. Compared with these methods, RDA provides some flexibility while aligning features.

\subsection{Denoising Training}

It is verified in \cite{li2022dn} that, the instability of bipartite graph matching in DETR could result in slow convergence and hence hinder the performance. The proposed DETR-based detection learns to refine the coarse object features and boxes iteratively, which can be simulated by the process of reconstructing noisy ground-truth labels and boxes. Denoising Training, as an auxiliary task of denoising labels and boxes, has fixed target assigning results, so it can mitigate the effect of matching instability and accelerate the convergence. Besides, to simulate predicting both positive and negative samples, both positive and negative noisy targets are generated for each ground-truth target, which provides a more reasonable optimization goal. Additionally, to transfer the Denoising Training to the oriented object detection, we also design the task of denoising angles for training procedure.

Given a ground truth $gt=(c_{gt}, b_{gt}, \theta_{gt})$, where $c_{gt}$ is the class, $b_{gt}=(x_{c}, y_{c}, w, h)$ is the bounding box, and $\theta_{gt}$ is the angle, the noisy ground truth $gt_{n}=(c_{n}, b_{n}, \theta_{n})$ is obtained as follows.

The noisy labels $c_{n}$ are generated by randomly selecting part of the ground-truth labels and overlaying the selected labels with arbitrary object labels at a ratio of $\alpha$. The target labels of positive samples are assigned with the ground-truth labels, while those of negative samples are assigned with the background category.

The noisy boxes $b_{n}$ are generated by moving the four boundaries of ground-truth boxes randomly. Specifically, the ground truth box $b_{gt}=(x_{c}, y_{c}, w, h)$ is convert into format $\hat{b}_{gt}=(x_{l}, y_{u}, x_{r}, y_{b})$, where $x_{l}$ and $x_{r}$ represent horizontal ordinates of left and right boundaries, respectively, and $y_{u}$ and $y_{b}$ represent vertical ordinates of upper and bottom boundaries, respectively. The negative noisy boxes should have larger noise scale than positive ones, because the farther proposals should predict negative samples \cite{zhang2023dino}. Hence, random noise offsets $\hat{\epsilon}=(\Delta x_{l}, \Delta y_{u}, \Delta x_{r}, \Delta y_{b})$ where $\Delta x_{l}, \Delta x_{r} \sim U(-\frac{\beta}{2}w, \frac{\beta}{2}w)$ and $\Delta y_{u}, \Delta y_{b} \sim U(-\frac{\beta}{2}h, \frac{\beta}{2}h)$, are generated for positive noisy boxes. While for negative ones, $\Delta x_{l}, \Delta x_{r} \sim \left(U(-\beta w, -\frac{\beta}{2}w) + U(\frac{\beta}{2}w, \beta w) \right) $ and $\Delta y_{u}, \Delta y_{b} \sim \left(U(-\beta h, -\frac{\beta}{2}h) + U(\frac{\beta}{2}h, \beta h) \right) $. The noisy boxes are calculated with $\hat{b_{n}}=\hat{b_{gt}} + \hat{\epsilon}$ and then converted into format $b_{n}=(x_{nc}, y_{nc}, w_{n}, h_{n})$ as initial box proposals. The decoder learns to denoise $b_{n}$ and reconstruct $b_{gt}$.

The noisy angles $\theta_{n}$ are generated by shifting the $\theta_{gt}$ with $\theta_{n}=f(\theta_{gt}+\Delta \theta)$, where $\Delta \theta \sim U(-\gamma \pi, \gamma \pi)$ are generated for positive noisy angles and $\gamma$ is the angle noise scale. While for negative ones, $\Delta \theta \sim U(-2\gamma \pi, 2\gamma \pi)$. $f()$ is a periodic function, 
thereby ensuring that the $\theta_{gt}+\Delta \theta$ remain within the defined range.

\subsection{Aspect Ratio Sensitive Matching and Loss}

After decoding the output embeddings from decoder, the predicted results will be matched with targets to count the loss during the training. Let $y$ denote ground truth set of oriented objects and $\hat{y}$ denote the $N$ predictions. Then we calculate the cost between these two sets and search for a permutation of $N$ elements $\sigma \in {{O}_{n}}$ with the lowest cost:
\begin{equation}
    \hat{\sigma }=\underset{\sigma \in {{O}_{n}}}{\mathop{\arg \min }}\,\sum\limits_{i}^{N}{{{L}_{match}}({{y}_{i}},{{{\hat{y}}}_{\sigma (i)}})} ,
\label{eq:match}
\end{equation}
where ${{L}_{match}}({{y}_{i}},{{\hat{y}}_{\sigma (i)}})$ is a pair-wise matching cost between ground truth ${{y}_{i}}$ and a prediction with index $\sigma (i)$, which takes into account the class prediction, angle prediction and the similarity of predicted and ground truth horizontal boxes, and we define it as follows:

\begin{figure}[!tb]
    \centering
    \begin{minipage}[t]{0.9\linewidth}
        \centering
        \includegraphics[width=0.98\linewidth]{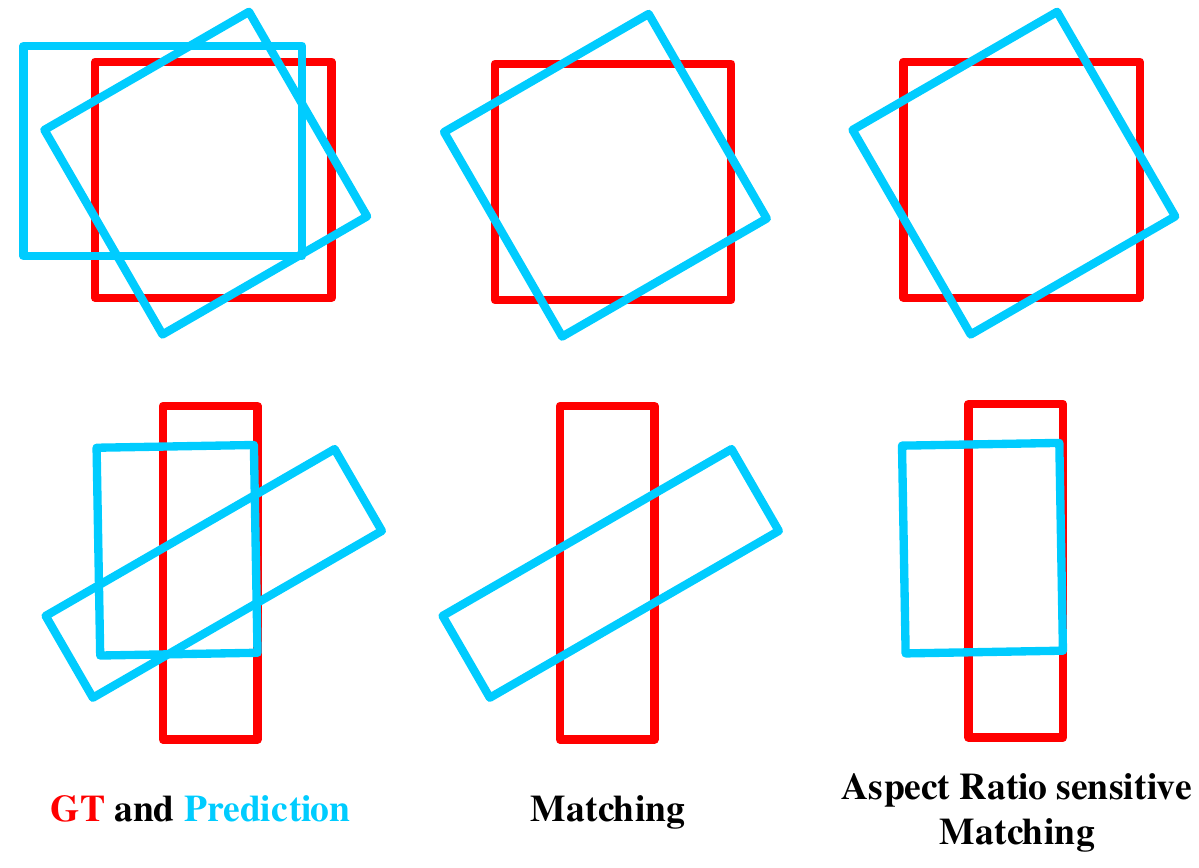}
    \end{minipage}%
    \caption{Illustration of Aspect Ratio sensitive Matching. ‘GT’ means ground truth.}
    \label{fig:ARAM}
\end{figure}

\begin{equation}
    \begin{aligned}
       {{L}_{match}}({{y}_{i}},{{{\hat{y}}}_{\sigma (i)}}) &={{\lambda }_{cls}}\cdot {{L}_{cls}}({{c}_{i}},{{{\hat{c}}}_{\sigma (i)}}) \\ 
     & +{{\lambda }_{box}}\cdot {{L}_{box}}({{b}_{i}},{{{\hat{b}}}_{\sigma (i)}}) \\ 
     & +{{\lambda }_{iou}}\cdot {{L}_{iou}}({{b}_{i}},{{{\hat{b}}}_{\sigma (i)}}) \\ 
     & +{{\lambda }_{\theta }}\cdot {{L}_{\theta }}({{\theta }_{i}},{{{\hat{\theta }}}_{\sigma (i)}}) ,
    \end{aligned}
    \label{eq:match_loss}
\end{equation}
where the $c$ is the class label, $b$ is the horizontal box, and $\theta $ is the angle, respectively. Additionally, the ${{L}_{cls}}$ is the focal loss, the ${{L}_{box}}$ is the L1 loss, the ${{L}_{iou}}$ is the generalized intersection over union loss (GIoU), and ${{L}_{\theta }}$ is the cross entropy loss.

Whereas, considering the objects with larger aspect ration are more sensitive to the angle, we introduce a dynamic coefficient to adjust the angle loss, named Aspect Ratio sensitive Matching (ARM), which can be formulated as follows:
\begin{equation}
\begin{aligned}
L({{\theta }_{i}},{{\hat{\theta }}_{\sigma (i)}})\to \frac{2{{k}_{i}}}{1+{{k}_{i}}}L({{\theta }_{i}},{{\hat{\theta }}_{\sigma (i)}}) ,
\end{aligned}
\label{eq:arw}
\end{equation}
where $k_{i}$ is the aspect ratio of the targets and $k_{i} \geq 1$.

As shown in Fig.~\ref{fig:ARAM}, the red ground truth will be matched to the blue prediction whose cost is the lowest according to Eq.~\ref{eq:match_loss}. When the ARM is introduced, the ground truth with a large aspect ratio will consider more about the angle deviation in the cost, so it is prone to be matched to the prediction whose angle is more similar with its.

The definition of training loss function is consistent with the Eq.~\ref{eq:match_loss}. Meanwhile, we also introduce the Aspect Ratio sensitive Loss (ARL), which is the same as Eq.~\ref{eq:arw}, to dynamically adjust the angle loss.

\section{Experiments} \label{sec:exp}
\subsection{Datasets and Implementation Details}
\textbf{DOTA-v1.0} \cite{xia2018dota} is one of the largest datasets for oriented object detection, containing 2,806 large aerial images from different sensors and platforms ranging from around 800 $\times$ 800 to 4,000 $\times$ 4,000 and 188,282 instances. It has 15 common categories: Plane (PL), Baseball diamond (BD), Bridge (BR), Ground track field (GTF), Small vehicle (SV), Large vehicle (LV), Ship (SH), Tennis court (TC), Basketball court (BC), Storage tank (ST), Soccer-ball field (SBF), Roundabout (RA), Harbor (HA), Swimming pool (SP), and Helicopter (HC). We use both training and validation sets for training, and the test set for testing. We divide the images into 1024 $\times$ 1024 sub-images with an overlap of 200 pixels. During the training, only the random horizontal, vertical, diagonal flipping is adopted to avoid over-fitting and no other tricks are utilized. The performance of the test set is evaluated on the official DOTA evaluation server.

\textbf{DIOR-R} \cite{cheng2022anchor} is an aerial image dataset annotated by oriented bounding boxes from the DIOR \cite{li2020object} dataset. There are 23,463 images and 192,518 instances in this dataset, containing 20 common categories: Airplane (APL), Airport (APO), Baseball Field (BF), Basketball Court (BC), Bridge (BR), Chimney (CH), Expressway Service Area (ESA), Expressway Toll Station (ETS), Dam (DAM), Golf Field (GF), Ground Track Field (GTF), Harbor (HA), Overpass (OP), Ship (SH), Stadium (STA), Storage Tank (STO), Tennis Court (TC), Train Station (TS), Vehicle (VE) and Windmill (WM). DIOR-R has a high variation of object size, both in spatial resolutions, and in the aspect of inter-class and intra-class size variability across objects. Different imaging conditions, weather, seasons, image quality are the major challenges of DIOR-R. We use both training and validation sets for training and the test set for testing.

\textbf{OHD-SJTU} \cite{yang2022on} is a public dataset for oriented object detection and object heading detection. It contains two different scale datasets, called OHD-SJTU-S and OHD-SJTU-L. OHD-SJTU-S collects 43 large scene images ranging from 10,000 $\times$ 10,000 to 16,000 $\times$ 16,000 and 4,125 instances in this dataset. In contrast, OHD-SJTU-L adds more categories and instances, containing six object categories and 113,435 instances. In line with previous work \cite{yang2021r3det,yang2022on} processing, we divide the images into 600 $\times$ 600 sub-images with an overlap of 150 pixels and scale them to 800 $\times$ 800.

All models in this paper are implemented by PyTorch~\cite{paszke2019pytorch} based framework MMRotate \cite{zhou2022mmrotate}, and trained with AdamW \cite{loshchilov2018decoupled} optimizer. The initial learning rate is 10$^{-4}$ with 2 images per mini-batch with ‘3x’ training schedule on DOTA-v1.0, DIOR-R and OHD-SJTU-L and ‘9x’ training schedule on OHD-SJTU-S respectively. In addition, we adopt learning rate warm-up for 500 iterations, and the learning rate is divided by 10 at decay step.

\subsection{Ablation Studies}

\begin{table}[!tb]
    \caption{Comparison of AR-CSL and CSL with different radius on DOTA-v1.0.
    }
    \begin{center}
        \resizebox{0.48\textwidth}{!}{
            \begin{tabular}{cc|cccc|c} 
            \hline\hline
            \multicolumn{2}{c|}{\multirow{2}{*}{Method}} & \multicolumn{4}{c|}{CSL} & \multirow{2}{*}{AR-CSL}\\
            \cline{3-6}
            & ~ & R=2 & R=4 & R=6 & R=8 & \\
            \hline
            \multirow{2}{*}{\makecell[c]{Deformable \\ DETR}} & \multicolumn{1}{|c|}{AP$_{50}$} & \textbf{72.57} & 72.24 & 72.15 & 72.10 & 72.38 \\
            \cline{2-7}
            & \multicolumn{1}{|c|}{AP$_{75}$} & 43.61 & 42.82 & 44.07 & 43.19 & \textbf{45.71} \\
            \hline\hline
            \end{tabular}
        }
    \end{center}
    \label{tab:csl_and_arcsl}
\end{table}

\begin{table}[!tb]
    \caption{
    Comparison of AR-CSL and CSL under different angle discrete granularity $\omega$ on DOTA-v1.0.
    }
    \begin{center}
        \resizebox{0.50\textwidth}{!}{
            \begin{tabular}{c|c|c|c|c|c|c|c} 
                \hline\hline
                Method & Granularity & AP$_{50}$ & AP$_{75}$ & Method & Granularity & AP$_{50}$ & AP$_{75}$\\ 
                \hline
                \multirow{4}{*}{CSL(R=2)} & 
                $\omega$=1 & \textbf{72.57} & 43.61 & \multirow{4}{*}{CSL(R=$\frac{6}{\sqrt{\omega } }$)} & 
                $\omega$=1 & 72.15 & 44.07 \\
                & $\omega$=6 & 72.22 & 42.51 & & $\omega$=6 & 71.70 & 43.02\\
                & $\omega$=15 & 71.28 & 39.93 & & $\omega$=15 & 70.45 & 40.59 \\
                & $\omega$=30 & 71.39 & 27.62 & & $\omega$=30 & 70.55 & 28.72 \\
                \hline
                \multirow{4}{*}{CSL(R=6)} & 
                $\omega$=1 & 72.15 & 44.07 & \multirow{4}{*}{AR-CSL} & 
                $\omega$=1 & 72.38 & \textbf{45.71} \\
                & $\omega$=6 & 71.39 & 42.73 & & $\omega$=6 & 72.44 & 44.32\\
                & $\omega$=15 & 72.06 & 33.13 & & $\omega$=15 & 71.90 & 41.52 \\
                & $\omega$=30 & 71.01 & 19.61 & & $\omega$=30 & 71.38 & 30.25 \\
                \hline\hline
            \end{tabular}
        }
    \end{center}
    \label{tab:angle_discrete_granularity_and_radius}
\end{table}

\begin{table}[!tb]
    \caption{Comparison of different detectors using CSL and AR-CSL on DOTA-v1.0.}
    \begin{center}
        \resizebox{0.40\textwidth}{!}{
            \begin{tabular}{c|c|c|c} 
                \hline\hline
                \multicolumn{2}{c|}{Method($\omega=6$)} & AP$_{50}$ & AP$_{75}$ \\ 
                \hline
                \multirow{2}{*}{\shortstack{Deformable \\ DETR(R-50)}} & CSL(R=6) & 71.39 & 42.73 \\
                & AR-CSL & \textbf{72.44} & \textbf{44.32} \\
                \hline
                \multirow{2}{*}{RetinaNet(R-50)} & CSL(R=6) & 67.63 & 38.64 \\
                
                & AR-CSL & \textbf{67.98} & \textbf{39.18} \\
                \hline
                \multirow{2}{*}{FCOS(R-50)} & CSL(R=6) & 70.60 & 38.42 \\
                
                & AR-CSL & \textbf{71.60} & \textbf{39.74} \\
                \hline\hline
            \end{tabular}
        }
    \end{center}
    \label{tab:different_detectors}
\end{table}

\subsubsection{Studies on AR-CSL}
In this subsection we adopt Deformable DETR as baseline and compare AR-CSL with CSL and other angle classification methods.

\begin{figure*}[!tb]
        \centering
        \includegraphics[width=1.0\linewidth]{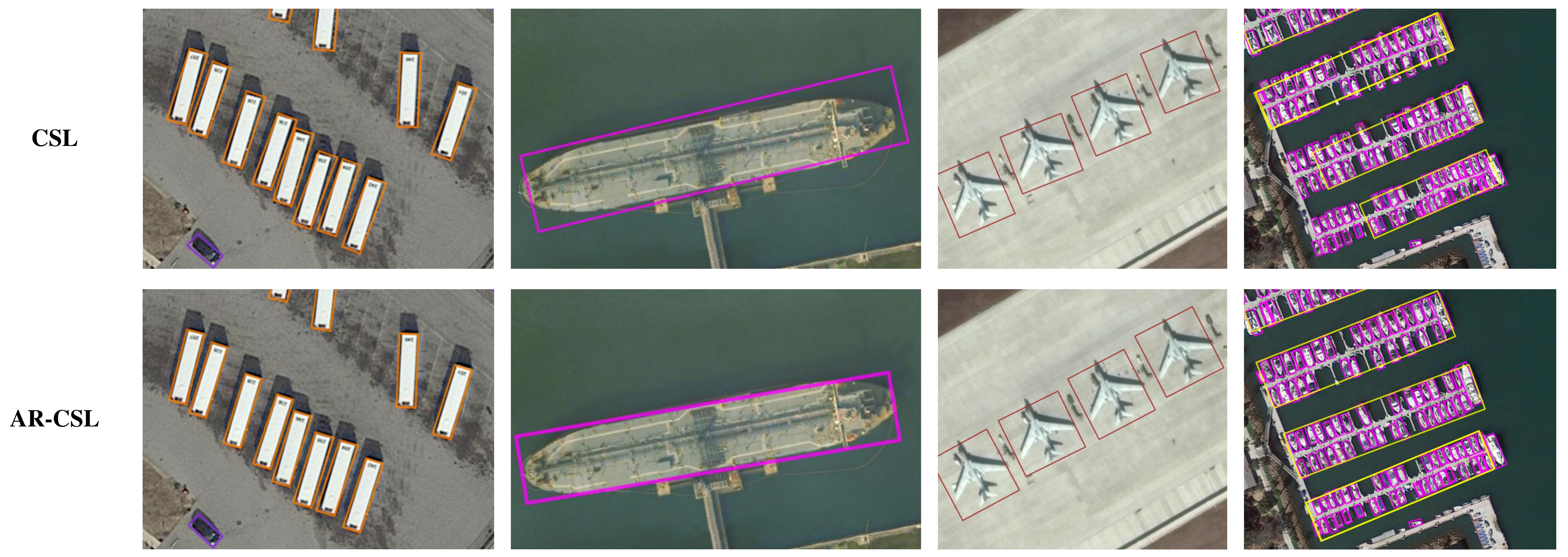}    
\caption{Visual comparison between CSL and AR-CSL on DOTA-v1.0.}
\label{fig:comparison}
\end{figure*}

\textbf{Comparison of CSL and AR-CSL on Deformable DETR.}
We compare the proposed AR-CSL and CSL with different radius $R$ under different angle discrete granularity $\omega$ on DOTA-v1.0, which is based on Deformable DETR, and the results are shown in Tab.~\ref{tab:csl_and_arcsl} and Tab.~\ref{tab:angle_discrete_granularity_and_radius}. Firstly, in Tab.~\ref{tab:csl_and_arcsl}, we set the $\omega$ to 1 and compare AR-CSL and CSL with different radius $R$. Fig.~\ref{fig:comparison} shows the visualization of the CSL and AR-CSL. The influence of $R$ on CSL is mainly concentrated on high-precision oriented detection and the maximum gap could reach 1.25\% on AP$_{75}$ ( $R=6$ with 44.07\% vs $R=4$ with 42.82\%). In contrast, AR-CSL achieves the best performance (about 45.71\% in terms of AP$_{75}$) without tuning any hyperparameters. Secondly, in Tab.~\ref{tab:angle_discrete_granularity_and_radius}, we set the $R$ in CSL to 2 and 6 and compare them with AR-CSL under different angle discrete granularity $\omega$. With the increase of the $\omega$, the angle interval becomes larger and the angle representation of each angle category becomes more ambiguous, thus the performances of CSL and AR-CSL on AP$_{75}$ deteriorate rapidly. However, the performance of CSL on AP$_{75}$ is more sensitive to the change of $\omega$, which needs to further tune the $R$. When the $\omega$ is 1 and 6, the best $R$ is 6 (44.07\% and 42.73\% on AP$_{75}$, respectively). When the $\omega$ is 15 and 30, the best $R$ is 2 (39.93\% and 27.62\% on AP$_{75}$, respectively). This is because the correlation among adjacent angle categories decreases with the increase of the $\omega$, thus a small $R$ could be more suitable. On the contrary, AR-CSL could dynamically smooth angle label with the change of $\omega$ so it is less affected by $\omega$ and performs better. 
Furthermore, we also modify the $R$ in CSL with $\frac{R}{\sqrt{\omega }}$ so that the radius can dynamically adjust itself under different angle discrete granularities to some extent. Compared with CSL with a fixed radius, the modified CSL has a better performance, but it still lags behind AR-CSL.

\begin{figure}[!tb]
    \begin{minipage}[t]{1.0\linewidth}
        \centering
        \includegraphics[width=1.0\linewidth]{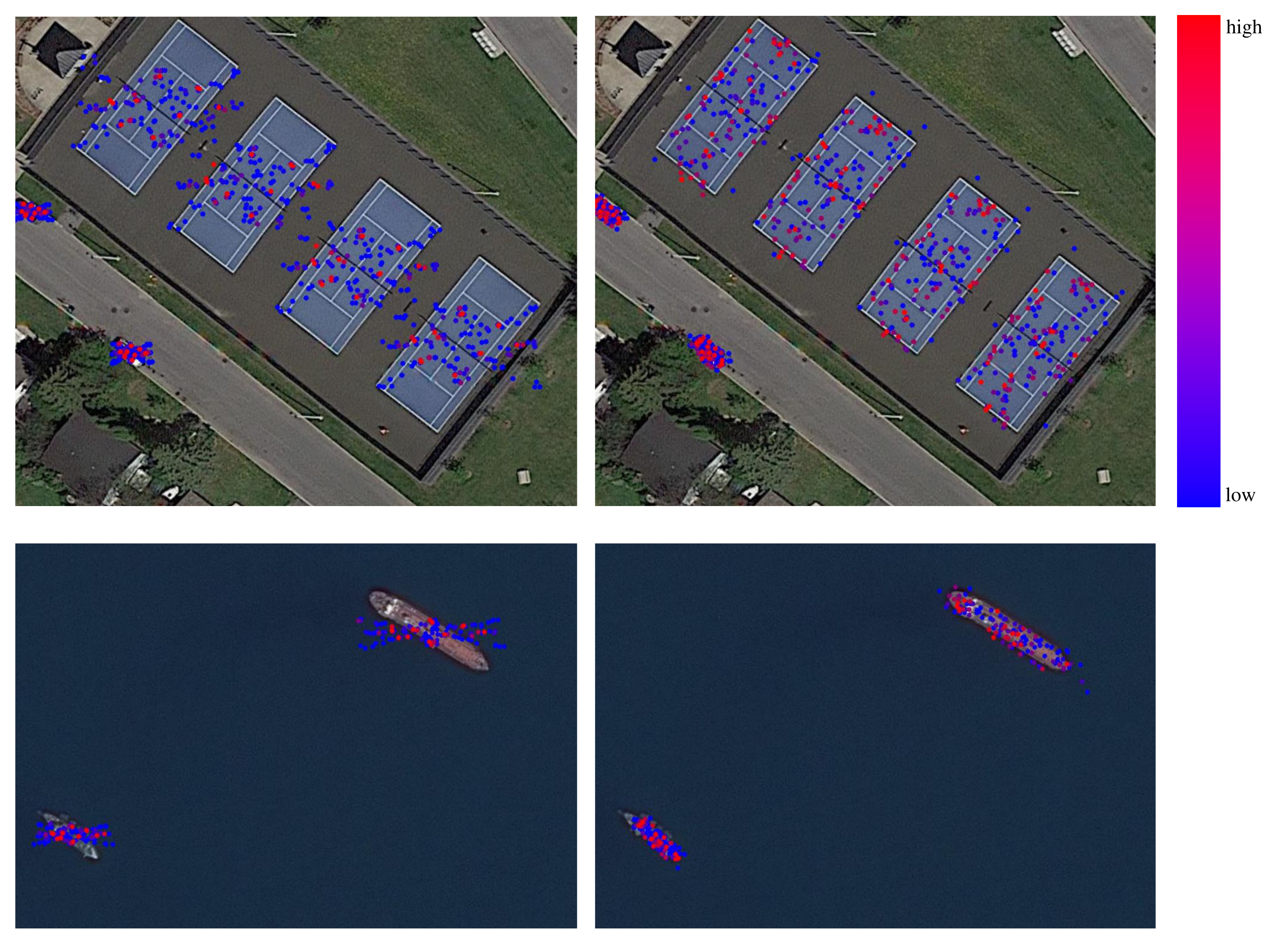}
    \end{minipage}%
\centering
    \caption{The location of sampling points before (left) and after (right) using RDA module. Each sampling point is marked as a filled circle whose color indicates its attention weight.}
    \vspace{-8pt}
    \label{fig:visualization_of_RDA}
\end{figure}

\textbf{Comparison of different detectors using CSL and AR-CSL.}
We conduct the CSL and AR-CSL on other detectors to verify the generalization of AR-CSL and set the $\omega$ to 6.
The experimental results are shown in Tab.~\ref{tab:different_detectors}. When the detector is changed to RetinaNet, AR-CSL achieves 67.98\% on AP$_{50}$ and 39.18\% on AP$_{75}$. When the detector is changed to FCOS, AR-CSL achieves 71.60\% on AP$_{50}$ and 39.74\% on AP$_{75}$. Compared with CSL, AR-CSL also performs well on AP$_{75}$ when using different detectors.

\begin{table}[!tb]
    \caption{Comparison of different angle classification methods on DOTA-v1.0.}
    \begin{center}
        \resizebox{0.48\textwidth}{!}{
            \begin{tabular}{cc|c|c|c|c|c} 
            \hline\hline
            \multicolumn{2}{c|}{Method} & Reg & \makecell[c]{CSL \\ \cite{yang2020arbitrary}} & \makecell[c]{POE \\ \cite{ming2023task}} & \makecell[c]{AR-BCL \\ \cite{xiao2023aspect}}  & AR-CSL\\
            \hline
            \multirow{2}{*}{\makecell[c]{Deformable \\ DETR}} & \multicolumn{1}{|c|}{AP$_{50}$} & 69.39 & 72.15 & 72.34 & 72.27 & \textbf{72.38} \\
            \cline{2-7}
            & \multicolumn{1}{|c|}{AP$_{75}$} & 40.79 & 44.07 & 43.75 & 44.67 & \textbf{45.71} \\
            \hline\hline
            \end{tabular}
        }
    \end{center}
    \label{tab:classification_method_compare}
\end{table}

\begin{table}[!tb]
    \caption{
    Ablation study of different angle prediction types and ways on DOTA-v1.0.
    }
    \begin{center}
        \resizebox{0.45\textwidth}{!}{
            \begin{tabular}{c|c|c|c} 
                \hline\hline
                Angle Pred. Type & Angle Pred. Way & AP$_{50}$ & AP$_{75}$ \\ 
                \hline
                \multirow{2}{*}{Regression} & $\theta = \theta_{ref} + \Delta \theta_{pred}$ & 69.48 & 40.32 \\
                & $\theta=\theta_{pred}$ & 69.39 & 40.79 \\ 
                \hline
                \multirow{2}{*}{\shortstack{Classification \\ (AR-CSL)}} & $\theta = \theta_{ref} + \Delta \theta_{pred}$ & 70.81 & 40.64 \\
                & $\theta=\theta_{pred}$ & \textbf{72.38} & \textbf{45.71} \\
                \hline\hline
            \end{tabular}
        }
    \end{center}
    \label{tab:angle_predict_form}
\end{table}

\begin{table}[!tb]
    \caption{Ablation study of ARS-DETR components on DOTA-v1.0.}
    \begin{center}
        \resizebox{0.48\textwidth}{!}{
            \begin{tabular}{c|cccc|cc} 
                \hline\hline
                DN & RDA & ARM & ARL & AN & AP$_{50}$  & AP$_{75}$ \\ 
                \hline
                ~ & ~ & ~ & ~ & ~ & 72.38 & 45.71 \\
                \checkmark & ~ & ~ & ~ & ~ & 73.14 \textbf{\small (+0.76)} & 47.04 \textbf{\small (+1.33)} \\
                \checkmark  & \checkmark & ~ & ~ & ~ & 72.80 \textbf{\small (+0.42)} & 48.06 \textbf{\small (+2.35)} \\
                \checkmark & ~ & \checkmark & ~ & ~ & 73.41 \textbf{\small (+1.03)} & 47.63 \textbf{\small (+1.92)} \\
                \checkmark & ~ & ~ & \checkmark & ~ & 73.31 \textbf{\small (+0.93)} & 47.58 \textbf{\small (+1.87)}  \\
                \checkmark & ~ & \checkmark & \checkmark & ~ & 73.43 \textbf{\small (+1.05)} & 48.13 \textbf{\small (+2.42)}  \\
                \checkmark & \checkmark & \checkmark & \checkmark & ~ & 73.90 \textbf{\small (+1.52)} & 48.62 \textbf{\small (+2.91)} \\
                \checkmark & \checkmark & \checkmark & \checkmark & \checkmark & \textbf{74.16} \textbf{\small (+1.78)} & \textbf{49.41} \textbf{\small (+3.70)} \\
                \hline\hline
            \end{tabular}
        }
    \end{center}
    \label{tab:components_ablation}
\end{table}

\begin{table}[!tb]
    \caption{
    Comparison of different sampling methods on DOTA-v1.0.
    }
    \begin{center}
        \resizebox{0.48\textwidth}{!}{
            \begin{tabular}{c|c|cc} 
                \hline\hline
                ~ & Methods & AP$_{50}$  & AP$_{75}$ \\ 
                \hline
                \multirow{3}{*}{ARS-DETR} & Deformable Offsets & 73.47 & 46.94 \\
                ~ & Fixed Offsets & 73.58 & 48.76 \\
                ~ & RDA & \textbf{74.16} & \textbf{49.41} \\
                \hline\hline
            \end{tabular}
        }
    \end{center}
    \label{tab:cmp_of_RDA}
\end{table}

\begin{table}[!tb]
    \caption{Ablation study of different angle noise scale $\gamma$ on DOTA-v1.0.}
    \begin{center}
        \resizebox{0.48\textwidth}{!}{
            \begin{tabular}{cc|c|c|c|c|c} 
            \hline\hline
            \multicolumn{2}{c|}{$\gamma$} & 0.00 & 0.01 & 0.02 & 0.05  & 0.07\\
            \hline
            \multirow{2}{*}{\makecell[c]{ARS-DETR}} & \multicolumn{1}{|c|}{AP$_{50}$} & 73.90 & 73.46 & 73.91 & \textbf{74.16} & 73.72\\
            \cline{2-7}
            & \multicolumn{1}{|c|}{AP$_{75}$} & 48.62 & 48.86 & 49.23 & \textbf{49.41} & 48.85 \\
            \hline\hline
            \end{tabular}
        }
    \end{center}
    \label{tab:angle_noise_ablation}
\end{table}

\textbf{Comparison with other angle classification methods on Deformable DETR.}
To explore the effectiveness of our proposed AR-CSL, we also conduct several experiments using different angle classification methods on Deformable DETR and the results are shown in Tab.~\ref{tab:classification_method_compare}.
Reg is the baseline that uses regression-based method to predict angle and it achieves 69.39\% and 40.79\% on AP$_{50}$ and AP$_{75}$ respectively.
CSL\cite{yang2020arbitrary} transfers the angle prediction to a classification task and utilizes the Gaussian window to smooth the angle label, greatly improving the performance of oriented object detection with 2.76\% on AP$_{50}$ and 3.28\% on AP$_{75}$. Recent POE\cite{ming2023task} adopts a n-ary codes to predict angle but its performance in high-precision oriented object detection is slightly inferior to CSL. AR-BCL\cite{xiao2023aspect} considers the square-like objects and introduces bi-directional angle to improve CSL, which further prompt 0.6\% on AP$_{75}$. Our AR-CSL dynamically consider the object with different aspect ratio and achieves 45.71\% on AP$_{75}$, which is best among the compared methods.

\begin{table*}[!tb]
    \renewcommand{\arraystretch}{1.25}
    \caption{
    Comparisons with the advanced oriented detectors on DOTA-v1.0. R-50 indicates ResNet50 \cite{he2017mask}. Swin-T indicates Swin-Transformer \cite{liu2021swin}. ReR-50 indicates ReResNet50 \cite{han2021align}. D-DETR means Deformable DETR \cite{zhu2021deformable}. * indicates that the model adopts the 1x training schedule(12 epochs). \textbf{Red} and \textbf{blue}: top two performances.
    }
    \begin{center}
        \resizebox{1.0\textwidth}{!}{
            \begin{tabular}{l|c|c|c|c|c|c|c|c|c|c|c|c|c|c|c|c|c|c|c} 
            \hline\hline
            Method & Backbone & PL & BD & BR & GTF & SV & LV & SH & TC & BC & ST & SBF & RA & HA & SP & HC & AP$_{50}$ & AP$_{75}$ & AP$_{50:95}$\\ 
            \hline
            AO2-DETR~\cite{dai2022ao2} & R-50 & 87.99 & 79.46 & 45.74 & 66.64 & 78.90 & 73.90 & 73.30 & 90.40 & 80.55 & 85.89 & 55.19 & 63.62 & 51.83 & 70.15 & 60.04 & 70.91 & 22.60 & 33.31\\
            Rotated D-DETR \cite{zhu2021deformable} & R-50 & 84.89 & 70.71 & 46.04 & 61.92 & 73.99 & 78.83 & 87.71 & 90.07 & 77.97 & 78.41 & 47.07 & 54.48 & 66.87 & 67.66 & 55.62 & 69.48 & 40.32 & 40.27\\
            Rotated FCOS*~\cite{tian2019fcos} & R-50 & 89.06 & 76.97 & 47.92 & 58.55 & 79.78 & 76.95 & 86.90 & 90.90 & 84.87 & 84.58 & 57.11 & 64.68 & 63.69 & 69.38 & 46.87 & 71.88 & 37.30 & 39.80 \\
            Rotated FCOS~\cite{tian2019fcos} & R-50 & 88.52 & 77.54 & 47.06 & 63.78 & 80.42 & 80.50 & 87.34 & 90.39 & 77.83 & 84.13 & 55.45 & 65.84 & 66.02 & 72.77 & 49.17 & 72.45 & 39.84 & 41.02\\
            S$^2$A-Net*~\cite{han2021align} & R-50 & 89.25 & 81.19 & 51.55 & 71.39 & 78.61 & 77.37 & 86.77 & 90.89 & 86.28 & 84.64 & 61.21 & 65.65 & 66.07 & 67.57 & 50.18 & 73.91 & 35.52 & 39.05 \\
            S$^2$A-Net~\cite{han2021align} & R-50 & 89.26 & 84.11 & 51.97 & 72.78 & 78.23 & 79.41 & 87.46 & 90.85 & 85.62 & 84.09 & 60.18 & 65.90 & 72.54 & 71.59 & 55.31 & 75.29 & 40.08 & 42.00\\
            Rotated RetinaNet*~\cite{lin2017focal} & R-50 & 89.64 & 82.56 & 38.43 & 69.83 & 77.39 & 62.74 & 77.24 & 90.68 & 83.79 & 82.04 & 59.91 & 64.83 & 57.37 & 64.76 & 45.56 & 69.79 & 37.69 & 39.64\\
            Rotated RetinaNet~\cite{lin2017focal} & R-50 & 87.33 & 78.91 & 46.45 & 69.81 & 67.72 & 62.34 & 73.59 & 90.85 & 82.79 & 79.37 & 59.62 & 61.89 & 65.01 & 67.76 & 44.95 & 69.23 & 40.96 & 40.38\\
            Gliding Vertex*~\cite{xu2020gliding} & R-50 & 89.20 & 75.92 & 51.31 & 69.56 & 78.11 & 75.63 & 86.87 & 90.90 & 85.40 & 84.77 & 53.36 & 66.65 & 66.31 & 69.99 & 54.39 & 73.22 & 37.47 & 39.52\\
            Gliding Vertex~\cite{xu2020gliding} & R-50 & 88.71 & 77.22 & 52.00 & 70.85 & 73.75 & 74.81 & 86.55 & 90.89 & 80.41 & 84.63 & 57.66 & 62.88 & 68.49 & 71.86 & 58.17 & 73.26 & 41.14 & 41.29\\
            H2RBox~\cite{yang2023h2rbox} & R-50 & 88.16 & 80.47 & 40.88 & 61.27 & 79.78 & 75.25 & 84.40 & 90.89 & 80.05 & 85.35 & 58.91 & 68.46 & 63.67 & 71.87 & 47.18 & 71.77 & 41.42 & 41.49 \\
            R$^3$Det*~\cite{yang2021r3det} & R-50 & 89.29 & 75.21 & 45.41 & 69.23 & 75.53 & 72.89 & 79.28 & 90.88 & 81.02 & 83.25 & 58.81 & 63.15 & 63.40 & 62.21 & 37.41 & 69.80 & 36.59 & 37.82\\
            R$^3$Det~\cite{yang2021r3det} & R-50 & 89.24 & 83.32 & 48.03 & 72.52 & 77.52 & 76.72 & 86.48 & 90.89 & 82.33 & 83.51 & 60.96 & 63.09 & 67.58 & 69.27 & 49.50 & 73.40 & 41.69 & 41.43\\
            KFIoU~\cite{yang2023kfiou} & R-50 & 89.20 & 76.40 & 51.64 & 70.15 & 78.31 & 76.43 & 87.10 & 90.88 & 81.68 & 82.22 & 64.65 & 64.84 & 66.77 & 70.68 & 49.52 & 73.37 & 42.71 & 41.70\\
            Rotated Faster RCNN*~\cite{ren2015faster} & R-50 & 89.25 & 82.44 & 50.05 & 69.34 & 78.17 & 73.59 & 85.91 & 90.89 & 84.08 & 85.50 & 57.66 & 60.96 & 66.25 & 69.22 & 57.74 & 73.40 & 39.61 & 40.75\\
            Rotated Faster RCNN~\cite{ren2015faster} & R-50 & 89.09 & 78.28 & 48.93 & 71.54 & 74.01 & 74.99 & 85.90 & 90.84 & 86.87 & 85.03 & 57.97 & 69.74 & 68.10 & 71.28 & 56.88 & 73.96 & 43.44 & 42.93\\
            Rotated D-DETR w/ CSL \cite{yang2020arbitrary} & R-50 & 86.27 & 76.66 & 46.64 & 65.29 & 76.80 & 76.32 & 87.74 & 90.77 & 79.38 & 82.36 & 54.00 & 61.47 & 66.05 & 70.46 & 61.97 & 72.15 & 44.07 & 42.72\\
            SASM~\cite{hou2022shape} & R-50 & 87.51 & 80.15 & 51.07 & 70.35 & 74.95 & 75.80 & 84.23 & 90.90 & 80.87 & 84.93 & 58.51 & 65.59 & 69.74 & 70.18 & 42.31 & 72.47 & 44.21 & 43.01\\
            KLD~\cite{yang2021learning} & R-50 & 89.08 & 84.18 & 43.77 & 72.33 & 79.85 & 73.58 & 85.69 & 90.88 & 85.14 & 81.96 & 65.86 & 64.60 & 63.60 & 68.26 & 53.19 & 73.46 & 44.74 & 43.70\\
            Rotated ATSS*~\cite{zhang2020bridging} & R-50 & 88.50 & 77.73 & 49.60 & 69.86 & 76.87 & 72.52 & 82.49 & 90.83 & 80.30 & 82.96 & 62.34 & 64.67 & 64.83 & 66.81 & 53.97 & 72.29 & 37.81 & 40.05 \\
            Rotated ATSS~\cite{zhang2020bridging} & R-50 & 88.94 & 79.89 & 48.71 & 70.74 & 75.80 & 74.02 & 84.14 & 90.89 & 83.19 & 84.05 & 60.48 & 65.06 & 66.74 & 70.14 & 57.78 & 73.37 & 44.95 & 43.53\\
            GWD*~\cite{yang2021rethinking} & R-50 & 88.92 & 77.03 & 45.90 & 69.30 & 72.53 & 64.06 & 76.40 & 90.87 & 79.20 & 80.45 & 57.68 & 64.37 & 63.60 & 64.74 & 48.26 & 69.55 & 38.91 & 39.50\\
            GWD~\cite{yang2021rethinking} & R-50 & 89.06 & 80.56 & 44.27 & 73.02 & 79.51 & 73.53 & 85.55 & 90.89 & 86.21 & 83.26 & 63.17 & 64.24 & 63.56 & 69.04 & 52.92 & 73.25 & 45.21 & 44.04\\
            PSC \cite{yu2023psc} & R-50 & 89.65 & 83.80 & 43.64 & 70.98 & 79.00 & 71.35 & 85.08 & 90.90 & 84.28 & 82.51 & 60.64 & 65.06 & 62.52 & 69.61 & 54.0 & 72.87 & 46.18 & 43.98 \\
            CFA~\cite{guo2021beyond} & R-50 & 88.34 & 83.09 & 51.92 & 72.23 & 79.95 & 78.68 & 87.25 & 90.90 & 85.38 & 85.71 & 59.63 & 63.05 & 73.33 & 70.36 & 47.86 & 74.51 & 46.55 & 44.41\\
            Oriented Reppoints*~\cite{li2022oriented} & R-50 & 87.78 & 77.67 & 49.54 & 66.46 & 78.51 & 73.11 & 86.58 & 90.86 & 83.75 & 84.34 & 53.14 & 65.63 & 63.70 & 68.71 & 45.91 & 71.71 & 41.39 & 40.88\\
            Oriented Reppoints~\cite{li2022oriented} & R-50 & 88.52 & 80.62 & 52.68 & 73.04 & 79.61 & 80.73 & 87.76 & 90.89 & 81.82 & 85.33 & 59.95 & 64.88 & 73.81 & 69.84 & 46.18 & 74.38 & 46.56 & 44.57\\
            RoI Trans.~\cite{ding2019learning} & R-50 & 88.70 & 83.66 & 54.65 & 72.72 & 73.77 & 78.05 & 87.39 & 90.90 & 80.64 & 84.76 & 60.73 & 63.98 & 77.61 & 73.32 & 54.48 & 75.03 & 48.86 & 45.84\\
            RoI Trans.~\cite{ding2019learning} & Swin-T & 88.44 & 85.53 & 54.56 & 74.55 & 73.43 & 78.39 & 87.64 & 90.88 & 87.23 & 87.11 & 64.25 & 63.27 & 77.93 & 74.10 & 60.03 & \textbf{76.49} & \textbf{50.15} & \textbf{47.60}\\
            \hline
            ARS-DETR & R-50 & 86.97 & 75.56 & 48.32 & 69.20 & 77.92 & 77.94 & 87.69 & 90.50 & 77.31 & 82.86 & 60.28 & 64.58 & 74.88 & 71.76 & 66.62 & 74.16 & 49.41 & 46.21\\
            ARS-DETR & Swin-T & 87.65 & 76.54 & 50.64 & 69.85 & 79.76 & 83.91 & 87.92 & 90.26 & 86.24 & 85.09 & 54.58 & 67.01 & 75.62 & 73.66 & 63.39 & \textbf{75.47} & \textbf{51.77} & \textbf{47.77}\\
            \hline\hline
            \end{tabular}
        }
    \end{center}
    \label{tab:dota}
\end{table*}

\begin{figure*}[!tb]
        \centering
        \includegraphics[width=1.0\linewidth]{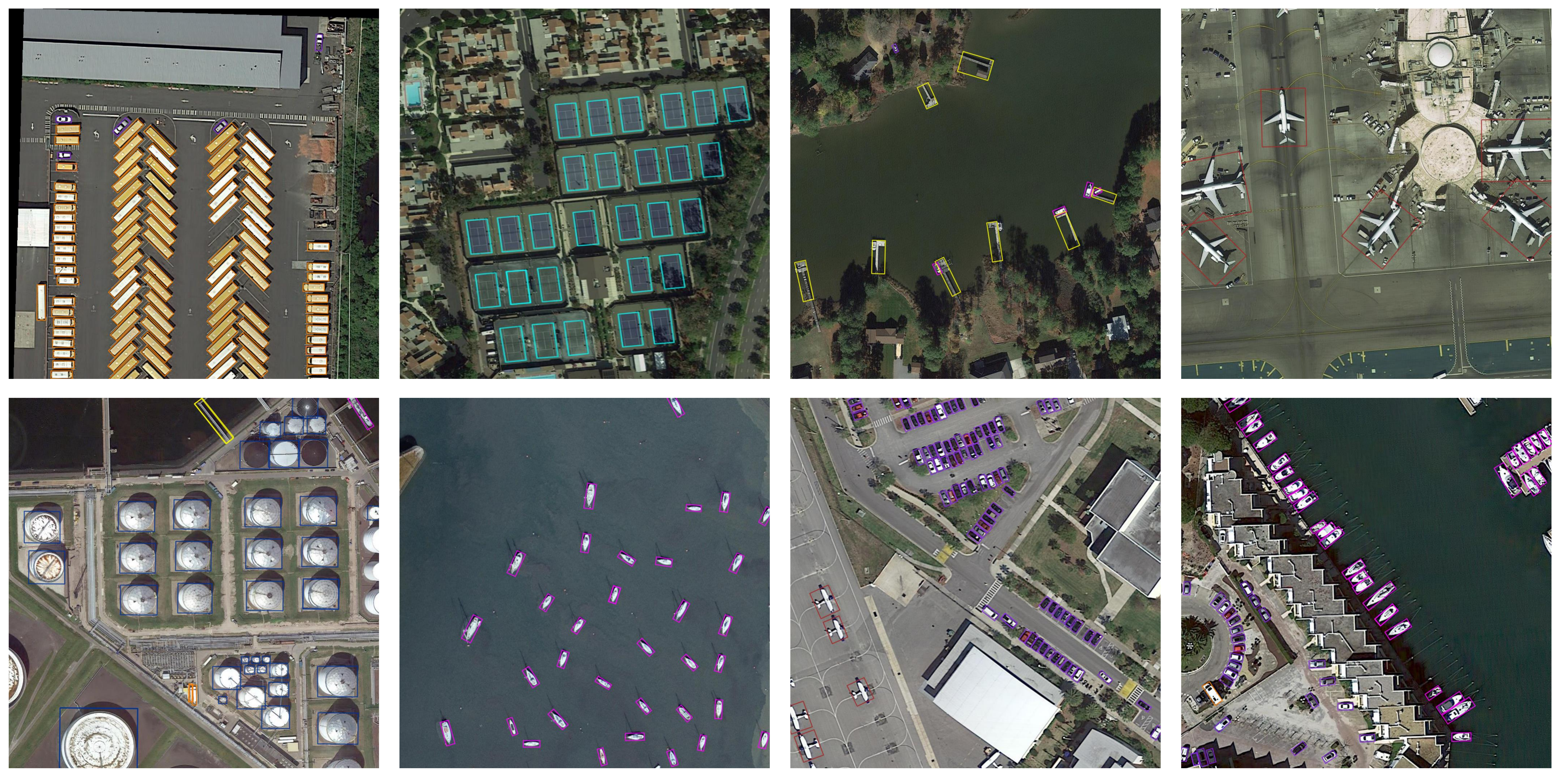}    
\caption{Examples of detection results on the DOTA-v1.0 dataset using ARS-DETR.}
\label{fig:visualization_dota}
\end{figure*}

\begin{table*}[!tb]
\renewcommand{\arraystretch}{1.3}
    \caption{Comparisons with the advanced oriented detectors on DIOR-R. \textbf{Red} and \textbf{blue}: top two performances.}
    \begin{center}
        \resizebox{1.0\textwidth}{!}{
            \begin{tabular}{l|c|c|c|c|c|c|c|c|c|c|c|c|c|c} 
            \hline\hline
            Method & \makecell[c]{Rotated \\ FCOS~ \\ \cite{tian2019fcos}} & \makecell[c]{S$^2$A-Net \\ ~\cite{han2021align}} & \makecell[c]{R$^3$Det \\ ~\cite{yang2021r3det}} & \makecell[c]{Gliding \\ Vertex~ \\ \cite{xu2020gliding}} & \makecell[c]{KFIoU~ \\ \cite{yang2023kfiou}} & \makecell[c]{SASM~ \\ \cite{hou2022shape}} & \makecell[c]{GWD~ \\ \cite{yang2021rethinking}} & \makecell[c]{KLD~ \\ \cite{yang2021learning}} & \makecell[c]{Rotated \\ Faster \\ RCNN~ \\ \cite{ren2015faster}} & \makecell[c]{Rotated \\ATSS~ \\ \cite{zhang2020bridging}} & \makecell[c]{CFA~ \\ \cite{guo2021beyond}} & \makecell[c]{Oriented \\ Reppoints~ \\ \cite{li2022oriented}} & \makecell[c]{RoI \\ Trans.~ \\ \cite{ding2019learning}}& ARS-DETR  \\ 
            \hline
            APL & 62.89 & 67.98 & 62.55 & 62.67 & 58.03 & 64.78 & 69.68 & 66.52 & 63.07 & 62.19 & 61.10 & 67.80 & 63.28 & 68.00 \\
            APO & 41.38 & 44.44 & 43.44 & 38.56 & 45.41 & 49.90 & 28.83 & 46.80 & 40.22 & 44.63 & 44.93 & 48.01 & 46.05 &  54.17 \\
            BF & 71.83 & 71.63 & 71.72 & 71.94 & 69.52 & 74.94 & 74.32 & 71.76 & 71.89 & 71.55 & 77.62 & 77.02 & 71.93 &  74.43 \\
            BC & 81.00 & 81.39 & 81.48 & 81.20 & 81.55 & 80.38 & 81.49 & 81.43 & 81.36 & 81.42 & 84.67 & 85.37 & 81.33 &  81.65 \\
            BR & 38.01 & 42.66 & 36.49 & 37.73 & 38.82 & 34.52 & 29.62 & 40.81 & 39.67 & 41.08 & 37.69 & 38.55 & 43.71 &  41.13 \\
            CH & 72.46 & 72.72 & 72.63 & 72.48 & 73.36 & 69.21 & 72.67 & 78.25 & 72.51 & 72.37 & 75.71 & 78.45 & 72.69 &  75.66 \\
            ESA & 77.73 & 79.03 & 79.50 & 78.62 & 78.08 & 76.28 & 76.45 & 79.23 & 79.19 & 78.54 & 82.68 & 81.13 & 80.17 & 81.92  \\
            ETS & 67.52 & 70.40 & 64.41 & 69.04 & 66.41 & 61.37 & 63.14 & 66.63 & 69.45 & 67.50 & 72.03 & 72.06 & 70.04 & 73.07  \\
            DAM & 28.61 & 27.08 & 27.02 & 22.81 & 25.23 & 31.66 & 27.13 & 29.01 & 26.00 & 30.56 & 33.41 & 33.67 & 31.42 &  34.89 \\
            GF & 74.58 & 75.56 & 77.36 & 77.89 & 79.24 & 72.22 & 77.19 & 78.68 & 77.93 & 75.69 & 77.25 & 76.00 & 78.00 &  76.10 \\
            GTF & 77.04 & 81.02 & 77.17 & 82.13 & 78.25 & 77.81 & 78.94 & 80.19 & 82.28 & 79.11 & 79.94 & 79.89 & 83.48 &  78.62 \\
            HA & 40.66 & 43.41 & 40.53 & 46.22 & 44.67 & 44.69 & 39.11 & 44.88 & 46.91 & 42.77 & 46.20 & 45.72 & 49.04 &  36.33 \\
            OP & 53.92 & 56.45 & 53.33 & 54.76 & 54.45 & 52.08 & 42.18 & 57.23 & 53.90 & 56.31 & 54.27 & 54.27 & 58.29 &  55.41 \\
            SH & 79.41 & 81.12 & 79.66 & 81.03 & 80.78 & 83.64 & 79.10 & 80.91 & 81.03 & 80.92 & 87.01 & 85.13 & 81.17 & 84.55 \\
            STA & 66.33 & 68.00 & 69.22 & 74.88 & 68.40 & 62.83 & 70.41 & 74.17 & 75.77 & 67.78 & 70.43 & 76.04 & 77.93 &  70.09 \\
            STO & 67.57 & 70.03 & 61.10 & 62.54 & 64.52 & 63.91 & 58.69 & 68.02 & 62.54 & 69.24 & 69.58 & 65.27 & 62.61 &  72.23 \\
            TC & 79.88 & 87.07 & 81.54 & 81.41 & 81.49 & 80.79 & 81.52 & 81.48 & 81.42 & 81.62 & 81.55 & 85.38 & 81.40 &  81.14 \\
            TS & 48.10 & 53.88 & 52.18 & 54.25 & 51.64 & 56.54 & 47.78 & 54.63 & 54.50 & 55.45 & 55.51 & 59.76 & 56.05 &  61.52 \\
            VE & 46.22 & 51.12 & 43.57 & 43.22 & 46.03 & 43.58 & 44.47 & 47.80 & 43.17 & 47.79 & 49.53 & 48.02 & 44.18 &  50.57 \\
            WM & 64.79 & 65.31 & 64.13 & 65.13 & 59.50 & 63.14 & 62.63 & 64.41 & 65.73 & 64.10 & 64.92 & 68.92 & 66.44 &  70.28 \\
            \hline
            AP$_{50}$ & 62.00 & 64.50 & 61.91 & 62.91 & 62.29 & 62.21 & 60.31 & 64.63 & 63.41 & 63.52 & 65.25 & \textbf{66.31} & 64.97 &  \textbf{66.12} \\
            AP$_{75}$ & 36.10 & 38.24 & 38.40 & 40.00 & 40.20 & 40.40 & 40.90 & 41.60 & 41.80 & 42.61 & 43.41 & 44.36 & \textbf{46.02} &  \textbf{45.81} \\
            AP$_{50:95}$ & 37.61 & 38.02 & 37.84 & 38.34 & 38.52 & 39.51 & 39.70 & 40.34 & 39.72 & 40.72 & 42.18 & 42.81 & \textbf{43.31} &  \textbf{43.89} \\
            \hline\hline
            \end{tabular}
        }
    \end{center}
    \vspace{-8pt}
    \label{tab:diorr}
\end{table*}

\subsubsection{Studies on ARS-DETR}
In this subsection, we adopt Deformable DETR + AR-CSL as our baseline and explore the effectiveness of our proposed methods.

\textbf{Ablation study of different angle prediction ways in DETR.}
The prediction types of angle mainly include regression and classification, and each type can be divided into ways in DETR: direct prediction ($\theta=\theta_{pred}$) or 
residual prediction
($\theta = \theta_{ref} + \Delta \theta_{pred}$). Tab. \ref{tab:angle_predict_form} compares the four combinations and finds that 
the way of direct prediction is the best in both regression and classification. 
We suspect that the periodicity of angle results in two optimization directions for the residual prediction during the optimization, which leads to the boundary problem and thus hinders the performance to some extent.

\textbf{Ablation study of Rotated Deformable Attention Module.}
To explore the effectiveness of the Rotated Deformable Attention Module (RDA), we perform ablation studies of detector w/ and w/o RDA in Tab. \ref{tab:components_ablation}. By aligning sampling points and features, RDA gets 1.02\% and 0.49\% gains from 47.04\% and 48.13\% to 48.06\% and 48.62\% on AP$_{75}$, respectively. The visualization is shown in Fig.~\ref{fig:visualization_of_RDA}. Before using RDA, there is a large number of sampling points will be distributed in the background around the objects or the adjacent objects, resulting in misalignment. Meanwhile, sampling points with high attention weight are relatively less and arranged densely. In contrast, after using RDA, sampling points could basically align with objects, and at the same time, sampling points with high attention weight are more widely distributed, indicating that the model pays more attention to multiple parts of objects.
Besides, as shown in Tab.~\ref{tab:cmp_of_RDA}, RDA also surpasses other two sampling methods on both AP$_{50}$ and AP$_{75}$, exhibiting its superiority.

\textbf{Ablation study of Aspect Ratio sensitive Matching and Loss.}
To investigate the contribution of Aspect Ratio sensitive Matching (ARM) and Aspect Ratio sensitive Loss (ARL), we conduct detailed ablation studies on Tab.~\ref{tab:components_ablation}. The results clearly show that when using ARM and ARL independently, they can improve the performance, about 0.59\% and 0.54\% on AP$_{75}$, respectively. When using both ARM and ARL, they can further improve the performance by 1.09\% on AP$_{75}$, verifying the effectiveness of ARM and ARL.

\textbf{Ablation study of Denoising training.}
During the experiments, we mainly explore the influence of the angle noise and separate it from the DN, indicating it with AN. Label noise $\alpha$ and box noise $\beta$ are set with 0.5 and 0.4 respectively, which is the same with DINO\cite{zhang2023dino}. Tab.~\ref{tab:components_ablation} 
shows that the basic DN training strategy (just adding noise to label and box) could improve the performance by 0.76\% and 1.33\% to 73.14\% and 47.04\% in terms of AP$_{50}$ and AP$_{75}$ respectively. Tab.\ref{tab:angle_noise_ablation} shows that when additionally adding noise to angle, it can also further improve the performance. When the $\gamma$ is set to 0.05, the best performance can be obtained with 74.16\% and 49.41\% on AP$_{50}$ and AP$_{75}$ respectively.
The results in Tab.~\ref{tab:components_ablation} and Tab.\ref{tab:angle_noise_ablation} show that noised ground truth could further help model learning for oriented object detection.

\begin{figure*}[!tb]
        \centering
        \includegraphics[width=1.00\linewidth]{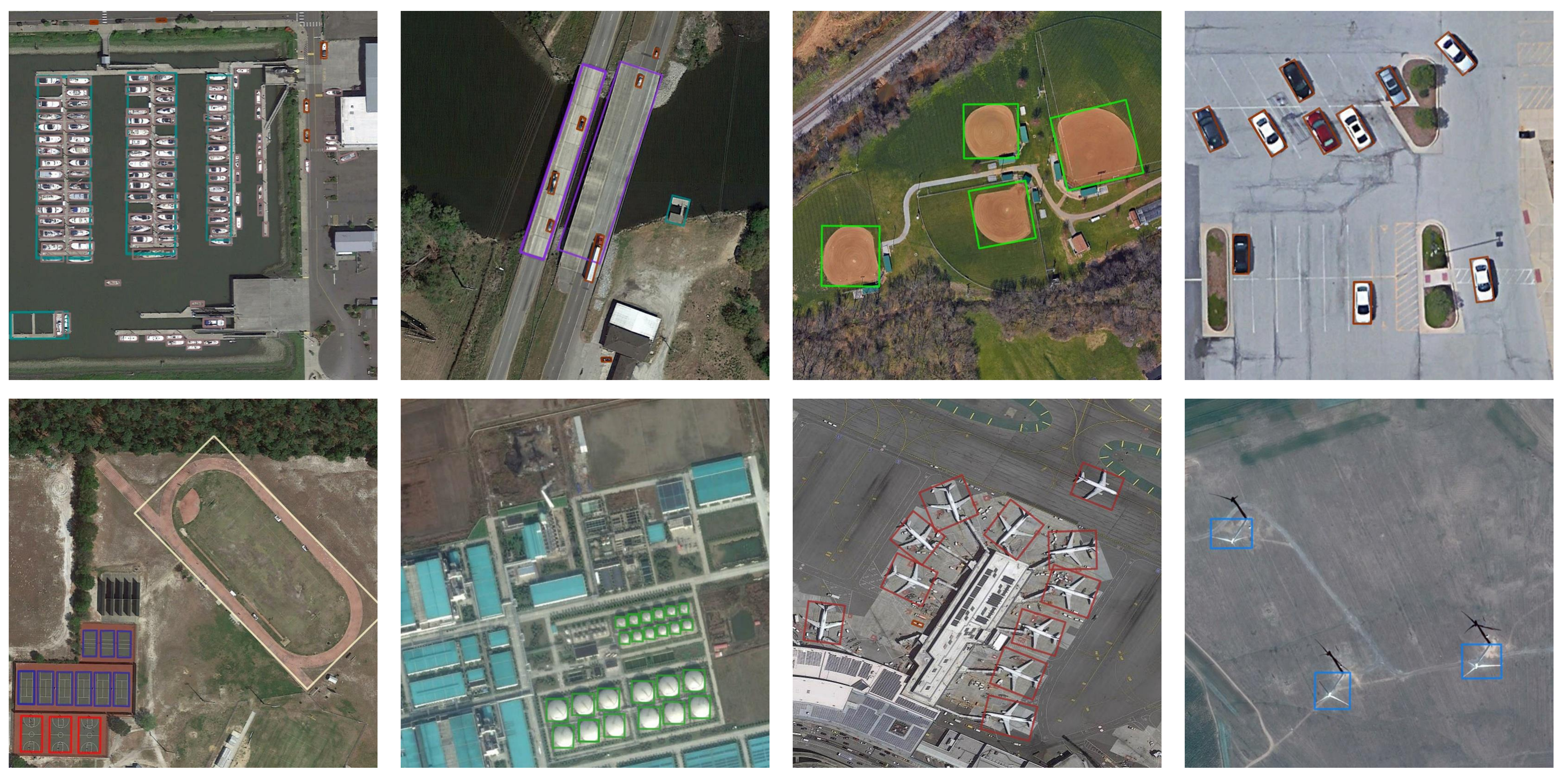}    
\caption{Examples of detection results on the DIOR-R dataset using ARS-DETR.}
\label{fig:visualization_dior}
\end{figure*}

\begin{table*}[!tb]
    \caption{Comparisons with the advanced oriented detectors on OHD-SJTU-L. \textbf{Red} and \textbf{blue}: top two performances.}
    \begin{center}
        \resizebox{0.95\textwidth}{!}{
            \begin{tabular}{l|c|c|c|c|c|c|c|c|c} 
            \hline\hline
            Method & PL & SH & SV & LV & HA & HC & AP$_{50}$ & AP$_{75}$ & AP$_{50:95}$  \\ 
            \hline
            RRPN~\cite{ma2018arbitrary} & 89.55 & 82.60 & 57.36 & 72.26 & 63.01 & 45.27 & 68.34 & 22.03 & 31.12    \\
            R$^2$CNN~\cite{jiang2017r2cnn} & 90.02 & 80.83 & 63.07 & 64.16 & 66.36 & 55.94 & 70.06 & 32.70 & 35.44    \\
            RetinaNet-H~\cite{yang2021r3det} & 90.22 & 80.04 & 63.32 & 63.49 & 63.73 & 53.77 & 69.10 & 35.90 & 36.89    \\
            R$^3$Det~\cite{yang2021r3det} & 89.89 & 87.69 & 65.20 & 78.95 & 57.06 & 53.50 & 72.05 & 36.51 & 38.57    \\
            Rotated Faster RCNN~\cite{ren2015faster} & 81.44 & 88.30 & 66.58 & 75.11 & 67.97 & 48.19 & 71.32 & 37.43 & 39.28 \\
            Rotated ATSS~\cite{zhang2020bridging} & 81.21 & 88.73 & 71.67 & 76.51 & 71.48 & 38.50 & 71.33 & 39.82 & 40.37 \\
            RetinaNet-R~\cite{yang2021r3det} & 90.00 & 86.90 & 63.24 & 86.90 & 62.85 & 52.35 & \textbf{72.78} & 40.13 & 40.58    \\
            OHDet~\cite{yang2022on} & 89.73 & 86.63 & 61.37 & 78.80 & 63.76 & 54.62 & 72.49 & \textbf{43.60} & \textbf{41.29}    \\ 
            \hline
            ARS-DETR & 87.48 & 87.76 & 65.23 & 78.51 & 69.43 & 56.89 & \textbf{74.20} & \textbf{46.08}  & \textbf{43.22}    \\
            \hline\hline
            \end{tabular}
        }
    \end{center}
    \vspace{-8pt}
    \label{tab:ohd_l}
\end{table*}

\begin{table}[!tb]
    \caption{Comparisons with the advanced oriented detectors on OHD-SJTU-S. \textbf{Red} and \textbf{blue}: top two performances.}
    \begin{center}
        \resizebox{0.45\textwidth}{!}{
            \begin{tabular}{l|c|c|c|c|c} 
                \hline\hline
                Method & PL & SH & AP$_{50}$ & AP$_{75}$ & AP$_{50:95}$ \\ 
                \hline
                RRPN~\cite{ma2018arbitrary} & 90.14 & 76.13 & 83.13 & 27.87 & 40.74 \\
                R$^2$CNN~\cite{jiang2017r2cnn} & 90.91 & 77.66 & 84.28 & 55.00 & 52.80 \\
                RetinaNet-H~\cite{yang2021r3det} & 90.86 & 66.32 & 78.59 & 58.45 & 53.07 \\
                R$^3$Det~\cite{yang2021r3det} & 90.82 & 85.59 & 88.21 & 67.13 & 56.19 \\
                Rotated Faster RCNN~\cite{ren2015faster} & 90.83 & 79.18 & 85.01 & 62.73 & 52.17 \\
                Rotated ATSS~\cite{zhang2020bridging} & 90.81 & 86.49 & 88.65 & 72.53 & 59.51 \\
                RetinaNet-R~\cite{yang2021r3det} & 90.82 & 85.59 & \textbf{89.48} & 74.62 & 61.86 \\
                OHDet~\cite{yang2022on} & 90.74 & 87.59 & 89.06 & \textbf{78.55} & \textbf{63.94}    \\ 
                \hline
                ARS-DETR & 90.18 & 89.71 & \textbf{89.95} & \textbf{80.67} & \textbf{65.49}    \\
                \hline\hline
            \end{tabular}
        }
    \end{center}
    \vspace{-8pt}
    \label{tab:ohd_s}
\end{table}

\subsection{Comparison with state-of-the-art methods}
\textbf{Results on DOTA-v1.0.} We report the results of 16 oriented object detectors in Tab.~\ref{tab:dota}. Since different methods use different image resolutions with different data pre-processing, data augmentation, backbone, training strategies, various tricks and etc. in the original papers, we implement all detectors on MMRotate~\cite{zhou2022mmrotate}, using the same setting to make the comparison as fair as possible. All the results are obtained by single-scale training and testing, and adopted the ‘1x’ (12epochs) or ‘3x’ (36 epochs) training schedule. With R-50 and Swin-T as the backbone, our method obtains 74.16\% and 75.47\% on AP$_{50}$, and 49.41\% and 51.77\% on AP$_{75}$, respectively, as shown in Fig.~\ref{fig:visualization_dota}. When using the ResNet50 as backbone, the performance of ARS-DETR under AP$_{50}$ is not as good as that of many advanced oriented detectors, but it has obvious advantages in high-precision detection and surpasses other advanced detectors on AP$_{75}$. Specially, ARS-DETR outperforms RoI Trans by 0.55\% (49.41\% VS 48.86\%), Oriented Reppoints by 2.85\% (49.41\% VS 46.56\%), CFA by 2.86\% (49.41\% VS 46.55\%), GWD by 4.2\% (49.41\% VS 45.21\%). In addition, there are also some detectors perform well on AP$_{50}$ but degenerate a lot on AP$_{75}$ (e.g. S$^2$A-Net, 75.29\% on AP$_{50}$ and 40.08\% on AP$_{75}$) and some detectors are relatively not good at AP$_{50}$ but has a favorable performance on AP$_{75}$ (e.g. PSC, 72.87\% on AP$_{50}$ and 46.18\% on AP$_{75}$), which further proves that it is not suitable to only use AP$_{50}$ to judge the performance of the detector. What’s more, ARS-DETR also exceeds DETR-based detectors like Rotated Deformable DETR and AO2-DETR.

\textbf{Results on DIOR-R.}
Results on DIOR-R dataset are shown in Tab.~\ref{tab:diorr}.
All methods adopt ‘3x’ training schedule and use R-50 as backbone. ARS-DETR achieves 66.12\% and 45.81\% on AP$_{50}$ and AP$_{75}$, respectively. The visualization is shown in Fig.~\ref{fig:visualization_dior}.

\textbf{Results on OHD-SJTU.}
We also compare the performance of some oriented object detection methods on OHD-SJTU, mainly include R2CNN, RRPN, RetinaNet, R$^3$Det, OHDet. Without any bells and whistles, our ARS-DETR achieves 46.08\% and 80.67\% on AP$_{75}$ in OHD-SJTU-L and OHD-SJTU-S respectively, surpassing other advanced oriented object detectors. The detailed results are shown in Tab.~\ref{tab:ohd_l} and Tab.~\ref{tab:ohd_s}.

Besides, from the above comparisons, it can be observed that AP$_{50}$ is not accurate enough to represent the performance of oriented object detectors, especially the performance of high-precision oriented detection.
In short, this paper mainly advocates the use of more stringent indicators (e.g. AP$_{75}$) to further study high-precision oriented object detector.

\section{Conclusion}

In this paper, we analyze the correlation between the angle and objects with different aspect ratios in detail
and identify the flaws of the current metric (i.e. AP$_{50}$) in high-precision oriented object detection. The metric AP$_{50}$, which is widely used, has a large tolerance for angle deviation, which cannot very accurately reflect the performance of the oriented object detectors. Therefore, using the more stringent metric AP$_{75}$ to measure the performance is more reasonable. 
Then, we design an oriented object detector named ARS-DETR and find that dynamically adjust the smoothing in angle classification, matching and loss calculation process in DETR based on sensitivity of objects with different aspect ratios to angle can effectively boost the performance. Additionally, aligning the features in DETR’s decoder and adopting the denoising training strategy could further improve the DETR to adapt to the oriented object detection.
Compared with other advanced oriented detectors, ARS-DETR achieves higher detection accuracy especially in the more stringent metric among the various datasets. 
Furthermore, we hope that this method will facilitate future work in high-precision oriented object detection and the application of DETR in oriented object detection.


\vfill

\end{document}